\documentclass[pmlr,twocolumn,10pt]{jmlr} % W&CP article

\usepackage{booktabs}
\usepackage{siunitx}
\usepackage{bbm}
\usepackage{multirow}
\usepackage{subcaption}

\theorembodyfont{\upshape}
\theoremheaderfont{\scshape}
\theorempostheader{:}
\theoremsep{\newline}
\jmlrvolume{LEAVE UNSET}
\jmlryear{2023}
\jmlrsubmitted{LEAVE UNSET}
\jmlrpublished{LEAVE UNSET}
\jmlrworkshop{Conference on Health, Inference, and Learning (CHIL) 2023} % W&CP title

\title[Interpretable Missing Values in Healthcare]{Missing Values and Imputation in Healthcare Data: \\ Can Interpretable Machine Learning Help?}

\author{%
    \Name{Zhi Chen}
    \Email{zhi.chen1@duke.edu} \\
    \addr Duke University, USA
    \AND
    \Name{Sarah Tan}
    \Email{ht395@cornell.edu} \\
    \addr Cornell University, USA
    \AND
    \Name{Urszula Chajewska}
    \Email{urszc@microsoft.com} \\
    \addr Microsoft Research, USA
    \AND
    \Name{Cynthia Rudin}
    \Email{cynthia@cs.duke.edu} \\
    \addr Duke University, USA
    \AND
    \Name{Rich Caruana}
    \Email{rcaruana@microsoft.com} \\
    \addr Microsoft Research, USA
 }

\begin{document}

\maketitle

\begin{abstract}
Missing values are a fundamental problem in data science. Many datasets have missing values that must be properly handled because the way missing values are treated can have large impact on the resulting machine learning model. In medical applications, the consequences may affect healthcare decisions. There are many methods in the literature for dealing with missing values, including state-of-the-art methods which often depend on black-box models for imputation. In this work, we show how recent advances in interpretable machine learning provide a new perspective for understanding and tackling the missing value problem. We propose methods based on high-accuracy glass-box Explainable Boosting Machines (EBMs) that can help users (1) gain new insights on missingness mechanisms and better understand the causes of missingness, and (2) detect -- or even alleviate -- potential risks introduced by imputation algorithms. Experiments on real-world medical datasets illustrate the effectiveness of the proposed methods.
\end{abstract}

\paragraph*{Data and Code Availability:}
% This initial paragraph is \textbf{mandatory}. Briefly state what data you
% use (including citations if appropriate) and whether the data are
% available to other researchers.\footnote{An example data availability
% statement: This paper uses the MIMIC-III dataset
% \citep{johnson2016mimic}, which is available on the PhysioNet repository
% \citep{johnson2016physionet}.}
% If you are not sharing code, you must explicitly state that you are not
% making your code available. If you are making your code available, then
% at the time of submission for review, please include your code as
% supplemental material or as a code repository link; in either case, your
% code must be anonymized. If your paper is accepted, then you should
% de-anonymize your code for the camera-ready version of the paper. \emph{If
% you do not include this data and code availability statement for your
% paper, or you provide code that is not anonymized at the time of
% submission, then your paper will be desk-rejected.} Your experiments later
% could refer to this initial data and code availability statement if it is
% helpful (e.g., to avoid restating what data you use).
This paper uses two publicly available datasets: MIMIC-II \citep{saeed2002mimic} and CDC Birth Cohort Linked Birth - Infant Death Data Files \citep{CDC:InfantLinkedDatasets}, and a proprietary pneumonia mortality prediction dataset \citep{cooper2005predicting}. The experiments leverage InterpretML open source \citep{nori2019interpretml} software package, and experiment code is provided in the supplementary materials.

\paragraph*{Institutional Review Board (IRB):}
The research does not require IRB approval.

\section{Introduction}

Missing values are ubiquitous in most datasets and have significant impact on machine learning models, as most machine learning models do not naturally handle missing values. While one could simply delete rows or columns as a preprocessing step, so the learning algorithm is only given observed, non-missing samples as inputs, such methods only work when the missingness ratio is small and the feature values are missing completely at random (MCAR). Deleting cases with non-MCAR missing values risks changing the data distribution, losing what might be valuable information contained in the missing cases.

% In other cases, such deletions may ignore information covered by the missingness itself.
% For example, in a medical dataset, doctors only record abnormal heart rate, so all patients whose heart rate falls in the regular range have a missing value in heart rate column (see more in Section \ref{}). If we simply delete samples with missing values, then we directly drop a group of important samples.

To avoid potential risks, systematic studies on understanding and handling missing values have been conducted in statistics and machine learning. Mechanisms of missingness have been studied and classified into three main categories, missing completely at random (MCAR), missing at random (MAR), and missing not at random (MNAR). 
% Urszula - cutting to save space
%See Section \ref{sec:background:missingtypes} for more details.
Different types of missingness have different solutions. For example, data cleaning and deletion methods like listwise and pairwise deletion are often used for MCAR \citep{rubin1976inference}. In the MAR scenario, numerous imputation methods have been proposed. These include simple techniques like imputing missing values with the mean or median, using a unique value to denote missing, and using advanced statistical and machine learning models to impute the missing values. State-of-the-art imputation methods include discriminative models like MICE \citep{van2011mice}, MissForest \citep{stekhoven2015missforest}, KNN Imputer \citep{troyanskaya2001missing}, and matrix completion \citep{mazumder2010spectral,yu2016temporal}, and generative models like 
% Expectation Maximization [CITE] \textcolor{red}{change this to a canonical bayesian paper} and 
deep generative models \citep{yoon2018gain}. Since most of these methods are based on black-box machine learning methods and the accuracy and behavior of the final model depends on the imputed values, it is difficult for users to recognize and understand the potential harms that might be introduced by these imputation methods.
% Broadly, they can be categorized into discriminative or generative methods [CITE], and range from simple techniques like imputing missing values with the mean or using a unique value to denote missing, to more complex processes such as matrix completion [see , or imputing with complex models like random forests [CITE MissForest] or neural nets. 

Recently developed interpretable machine learning methods have been shown to be useful for debugging models and detecting issues with datasets \citep{adebayo2020debugging,koh2017understanding}. Interpretable machine learning methods provide a new opportunity to study missing values and revisit some of the classical methods for handling missing values. In this paper, we propose novel methods based on the Explainable Boosting Machine (EBM) \citep{lou2012intelligible, lou2013accurate, nori2019interpretml}, a high-accuracy, fully-interpretable glass-box machine learning method, to answer the following questions: (1) how interpretability can help users gain insights on the causes of missingness, and (2) how interpretability can help detect and avoid potential risks introduced by different imputation methods. We show that the glass-box models provide new insights into missingness mechanisms, and in some settings, suggest alternate ways of handling missing values, as well as new tools that can alert users when imputation can lead to unexpected problems.

\section{Related Work}
\label{sec:relatedwork}
Issues with missing value imputation methods, whether generative or discriminative, have been pointed out in the literature \citep{harel2007multiple, jelivcic2009use, ibrahim2012missing, li2015multiple, van2018flexible, sidi2018treatment}. For example, generative imputation methods have been criticized for placing assumptions on the underlying data distribution, not all of which are testable \citep{yoon2018gain}. \citet{waljee2013comparison} studied four discriminative imputation methods -- MissForest, mean imputation, nearest neighbor imputation, and multivariate imputation by
chained equations (MICE) -- on medical datasets modified to have missing completely at random (MCAR) values, finding that MissForest had the least imputation error for both continuous and categorical variables. Our paper shows that MissForest, despite its popularity, presents issues that practitioners should notice. 

Connections between missing value imputation and causal inference methods have been drawn. \citet{ding2018causal} pointed out that the unconfoundedness assumption in causal inference is similar to the missing at random (MAR) assumption in missing data analysis, with both fields relying on these untestable, yet critical assumptions. The interpretability techniques we use in this paper can be applied to datasets even if they have missing not at random (MNAR) values. This flexibility presents a contribution given how difficult it is to distinguish between MAR and MNAR in practice \citep{van2018flexible}.

Our work is related to recent work using explainability techniques to detect issues with datasets. \citet{adebayo2020debugging} investigate the ability of feature attribution methods to detect spurious correlations and mislabeled examples.  \citet{koh2017understanding} used influence functions applied to black-box models to detect mislabeled examples in data. Our work does not use black-box models, and focuses on debugging missing values, a key issue in many datasets. %https://link.springer.com/chapter/10.1007/978-3-030-75015-2_14, https://arxiv.org/pdf/2105.04505.pdf

Some AutoML tools perform automatic data cleaning. Both the Automatic Statistician   \citep{steinruecken2019automatic} and AlphaClean \citep{krishnan2019alphaclean} attempt to automatically impute missing values. Unlike these papers, our focus is not on fixing datasets automatically, but on helping users detect, understand and mitigate missing values problems.

% An application of the field of AutoML (see \cite{he2021automl} for a review) is automating data cleaning. For example, the Automatic Statistician   \cite{steinruecken2019automatic} creates summaries of datasets with minimal
% human intervention, while AlphaClean \cite{krishnan2019alphaclean} uses black-box bayesian optimization to learn the most optimal steps to clean a dataset. Unlike these papers, our focus is not on fixing issues with datasets automatically, but on surfacing dataset issues to a practitioner who decides how to act on the issues, hence our focus on using interpretable, rather than black-box models, that a practitioner can understand.

\section{Background}
\label{sec:background}
\subsection{Types of Missing Values}
\label{sec:background:missingtypes}
\cite{rubin1976inference} classified missingness mechanisms into three types: (1) Missing Completely At Random (MCAR): the missingness is unrelated to the data, i.e. the probability of missing is the same for all samples; (2) Missing At Random (MAR): in addition to complete randomness, the probability of missingness of a feature is determined from the observed values of the other features (3) Missing Not At Random (MNAR): the probability of missingness is also related to unobserved values in the data, e.g., the missingness is also related to the feature value itself.
\subsection{Missing Value Imputation}
Here, we describe the advanced imputation methods we investigate in this paper: MissForest 
\citep{stekhoven2015missforest} and KNN Imputation \citep{troyanskaya2001missing}.

The MissForest algorithm first makes an initial guess for the missing values using mean and mode imputation. Then it sorts the features according to the missing rate, and fits a random forest iteratively to predict and impute each missing feature from the other features until the imputed values converge. MissForest is a popular imputation method as it is capable of capturing non-linear and interaction effects between features to improve imputation accuracy, and can be applied to mixed data types (continuous and discrete). Note that, the framework of MissForest is similar to that of MICE \citep{van2011mice} --- the only difference is MissForest uses random forest while MICE uses linear model as base model for imputation.

KNN imputation imputes the missing values by the mean value of its K nearest neighbors in the training set. The distance of two samples is measured on the non-missing features in both samples. KNN imputation is fast and accurate but requires choosing a good distance metric and tuning the hyperparameter K.

\subsection{Explainable Boosting Machines}
\label{sec:ebm}
The methods proposed in this work are based on one interpretable machine learning model, the Explainable Boosting Machine (EBM). 
% Urszula - cutting to save space
%We provide a brief introduction to EBMs in this section.

\begin{figure}
    \centering
    \includegraphics[width=.95\linewidth]{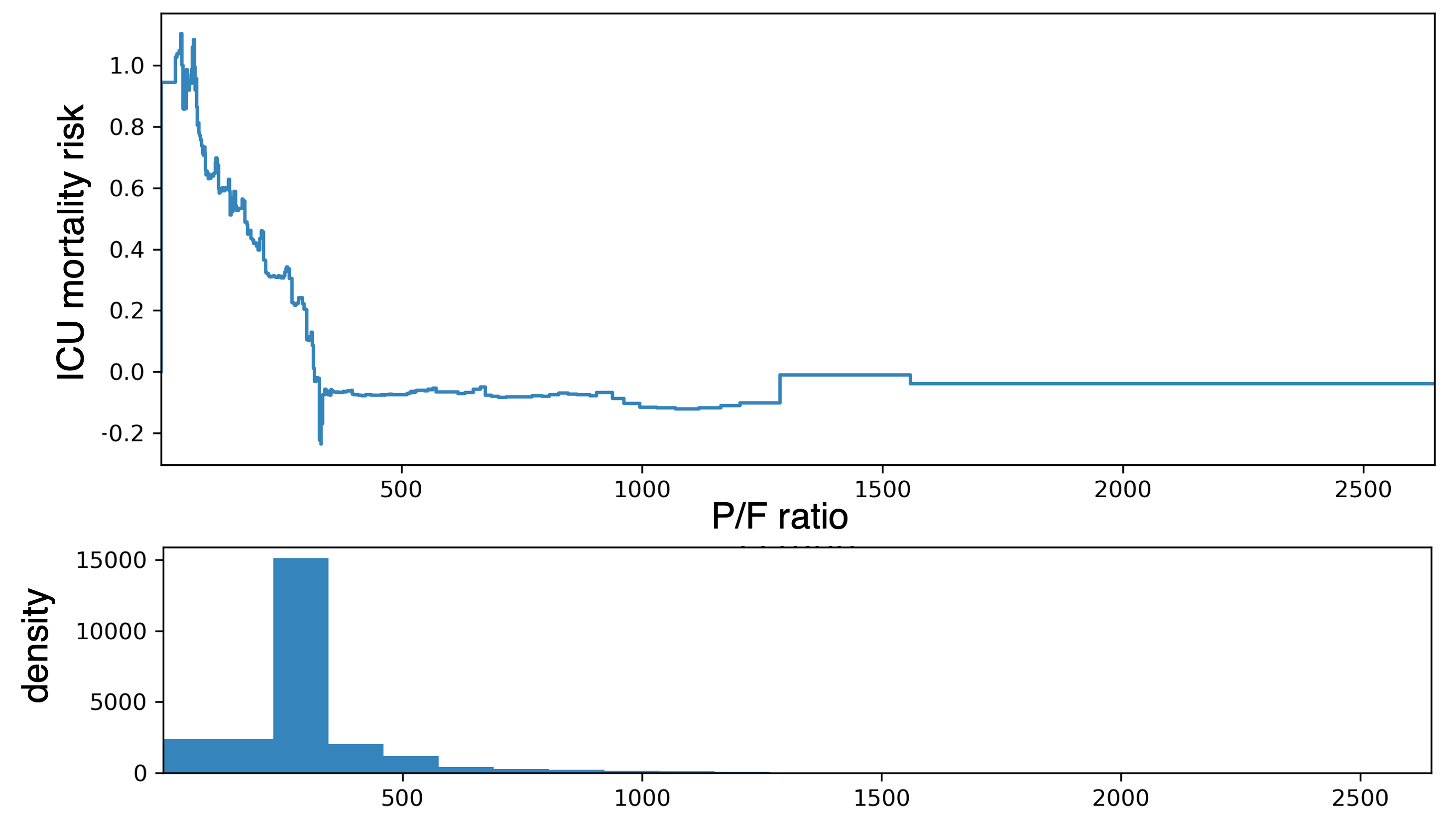}
    \caption{EBM shape function and density plot for P/F ratio when predicting ICU mortality.}
    \label{fig:missing_value_2}
\end{figure}

Suppose an input sample is denoted as $(\mathbf{x},y)$, where $\mathbf{x}$ is the $p$ dimensional feature vector and $y$ is the target. Denote the $j^{th}$ dimension of the feature vector as $x_j$. Then a generalized additive model (GAM), first introduced by \cite{hastie1987generalized}, is defined as
\begin{equation}
    g(E[y]) = \beta_0 + f_1(x_1) + f_2(x_2) + \cdots + f_p(x_p)
\end{equation}
where $\beta_0$ is the intercept, $f_j's$ are the shape functions and $g$ is the link function, e.g., the identity function for regression, or the logistic function for classification. Since one can add any offset to $f_j$ while subtracting it from $\beta_0$ or other shape functions, shape functions are often \textit{centered} by setting the population mean of $f_j$, i.e., $E_{x\sim \mathcal{X}}[f_j(x_j)]$ to 0. Because each shape function $f_j$ operates only on one single feature $x_j$, shape functions can be plotted. This makes GAMs interpretable since the model can be visualized as 2D graphs. In early work on GAMs, shape functions were often modeled as splines with smoothness constraints. Explainable Boosting Machines (EBMs) \citep{lou2012intelligible, lou2013accurate, nori2019interpretml} use bagged ensembles of boosted depth-restricted tree to represent each $f_j$. Tree-based ensemble learning significantly improves the performance of GAMs: EBMs outperform traditional GAMs because its shape functions have more representational power and better capture fine detail. Figure~\ref{fig:missing_value_2} shows the shape plot learned for P/F ratio (a measure of blood oxygenation) on the MIMIC II ICU mortality-risk classification problem.  The vertical axis is the contribution to risk on log scale: patients with low P/F ratio have high risk, and patients with P/F ratio near 1000 are low risk.  
%The low-risk spike at ratio=323 is due to missing values imputed with the mean.
EBM can further improves its accuracy by adding a small number of pairwise interactions, i.e.,
\begin{equation}
    g(E[y]) = \beta_0 + \sum_{j=1}^{p} f_j(x_j) + \sum_{k=1}^{K} f_k(x_{k_1}, x_{k_2}).
\end{equation}
Including pairwise interactions does not sacrifice interpretability since $f_k(x_{k_1}, x_{k_2})$ can be visualized as heatmaps. In this paper, we use EBM implemented in the InterpretML package \citep{nori2019interpretml}.

% \subsection{Types of Missingness}

% \subsection{Imputation Methods}

% \section{Missing Values}
% \label{sec:missing}
% A variety of problems can arise when there are missing values in data sets.  In this section, we explore a few of these issues, and show how interpretable models such as EBMs can be used to detect, and in some cases fix these problems.

\section{Gaining new insights on the causes of missingness}
\label{sec:missing}
\subsection{Missing Completely at Random}
\label{sec:missing:random}

When dealing with missing values, it is important to determine the mechanism of missingness. Standard statistical tests exist for testing MCAR, e.g., Little's test \citep{little1988test}. In this section, we propose a method to test MCAR based on EBM shape functions. The testing process of the proposed method can be directly visualized on the shape function plots, which is not achievable by Little's test. We will also show that EBM can bring additional insights beyond simply testing for MCAR.

\subsubsection{Testing for MCAR with EBM}
\label{sec:mcar}
To test for MCAR, we use the common trick of encoding  missing values with a unique value for the feature, e.g., -1 for a feature with positive values or a separate category for a categorical feature. After fitting an EBM that predicts the target, we get a shape function representing the contribution of different feature values for predicting the target, including the unique value denoting missingness. Note that the leaf nodes in EBM split the feature values into many bins, where each bin has a prediction score. These bins and scores together form the shape function. Therefore, the EBM shape function $f_j(\cdot)$ of feature $j$ can be rewritten as a linear combination of a series of indicator variables denoting if the feature values are within the bins, and the coefficients are the corresponding scores of the bins, i.e.,

\begin{equation}
    f_j(x_j) = \sum_{k=0}^{B_j-1}\theta_{j,k}\cdot \mathbbm{1}\{b_{j,k}<x_j<=b_{j,(k+1)}\},
\end{equation}
where $\{b_{j,k}\}_{k=0}^{B_j}$ are the bin edges of feature $j$ in the EBM model, and $\theta_{j,k}$ is the shape function score of the bin $(b_{j,k},b_{j,(k+1)}]$. Since EBM also uses the logistic link function, this transformation can turn EBM into a logistic regression model for binary classification. To do a statistical test, we need to make some assumptions and create a null hypothesis. First, we know that if the missingness is MCAR, i.e., all samples are missing with the same probability, the expected score for bins representing the missing value should be the same as the entire population, which is 0 as the shape function scores are mean centered in EBMs. Therefore, we can directly apply the classical significance Wald test of logistic regression coefficients \citep{kleinbaum2002logistic}. Specifically, our null hypothesis is $H_0:\ \theta_{i,k}=0$, and the alternative hypothesis is $H_1:\ \theta_{i,k}\neq0$. Then we calculate the square root of the Wald statistic
\begin{equation}\sqrt{W} = \frac{\hat{\theta}_{j,k}}{SE(\hat{\theta}_{j,k})},\end{equation}
find the $p$-value by assuming $\sqrt{W}$ follows a $Z$ distribution, and reject the null hypothesis if the p-value is smaller than a predefined threshold.
% Thus, we can directly do a t-test on the difference between the shape function score of the missing group and the average shape function score for the observed group, where the null hypothesis is the two groups having the same score. Note that, because we only test the difference between the scores, there exists the possibility that the two groups are significantly different but have the same score. However, in this case missingness does not affect the prediction.
% \textcolor{red}{Adding some formulas or results of t-test? This is actually more complicated than comparing the missing bar with 0, since t-test also requires to calculate the variance}
\begin{figure}
    \centering
    \includegraphics[width=.95\linewidth]{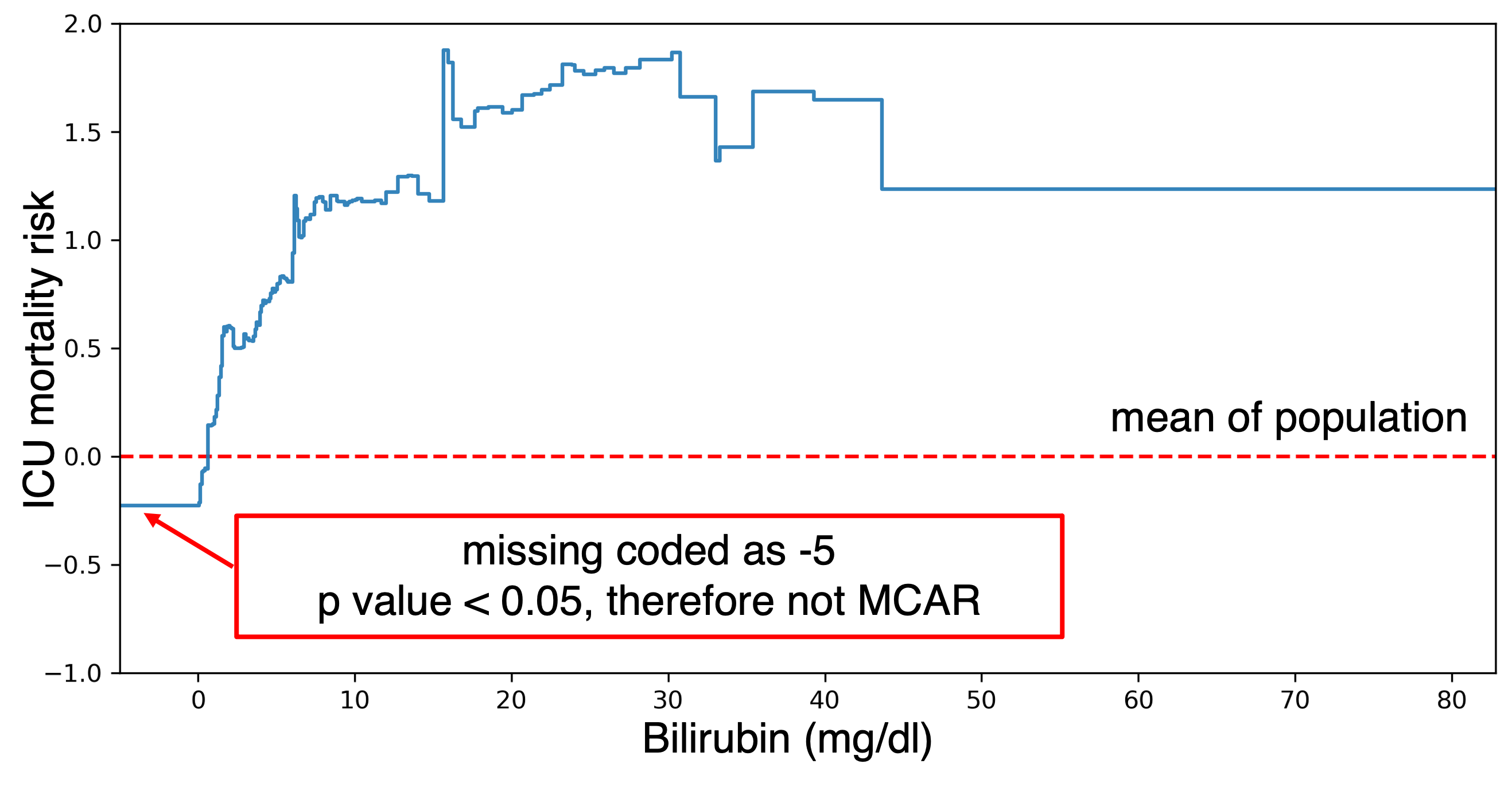}
    \caption{Example of using EBM shape function to test for MCAR. Missing coded as -5. $p$-value for testing MCAR is less than 0.05.}
    \label{fig:test_mcar}
\end{figure}
Figure \ref{fig:test_mcar} shows an example of using the proposed test for MCAR using EBM shape functions.
% \note{this is the first EBM plot, a clinician who hasn't seen this before might have trouble interpreting it. Section 3.3 might be more interesting with a plot and explanation of how to read it.} 
The missing value is encoded as -5 (lower than minimum possible feature value) for the Bilirubin feature.
% \note{Explain why we picked -5, very out of range of the feature?}
The Wald test rejects the null hypothesis and suggests that the missing value is not MCAR. 
% Using interpretable models such as EBMs allows us to visualize different types of missingness and test for properties such as not MCAR.
\begin{table}[h]
\centering
\scriptsize
\begin{tabular}{|c|c|c|c|c|c|c|}
\hline
Type & \multicolumn{3}{c|}{MCAR datasets$\downarrow$}  & \multicolumn{3}{c|}{MAR datasets$\uparrow$}      \\ \hline $p_{m}$ & 0.1 & 0.2 & 0.3 & 0.1 & 0.2 & 0.3 \\ \hline
Little's  &  \textbf{0.035} & 0.070 & 0.055 & \textbf{1.000} &  \textbf{1.000} & \textbf{1.000} \\ \hline
Ours  & 0.080 & \textbf{0.005} & \textbf{0.005} & 0.910  & 0.885  & 0.890\\ \hline
\end{tabular}
\caption{Proportion of times the test rejects the null hypothesis, i.e., the missing mechanism is MCAR, on datasets generated by different missing types, with different missing ratios $p_m$. When the data is truly MCAR, for which low rejection rate is desired, our method is less likely to reject the null hypothesis compared with Little's test, especially when missing ratio becomes larger. When the data is MAR, where we hope to reject the null hypothesis (high rejection rate is better), Little's test can reject all null hypothesis, and our method is able to reject it in most cases. \label{tab:mcar}}
\end{table}
% \note{what does conservative mean here? I'm also confused by the 10\% and 20\% and up and down arrows, explain in the caption?}

Table \ref{tab:mcar} compares the performance of the MCAR test we proposed with \citet{little1988test}. To test their performances, we generate semi-synthetic datasets where we know the ground truth missing mechanism. Specifically, we start from MIMIC-II dataset imputed by MissForest, and then add missing values to the ``Age'' feature manually\footnote{The ``Age'' feature has no missing values in the original dataset.}. For MCAR case, each sample has a fixed probability $p_{m}$ of missing the ``Age'' feature. For MAR case, we apply a linear model on all features except ``Age'', whose coefficients are randomly sampled from standard normal distribution to all samples in the dataset, adding a standard Gaussian noise to the output score, and the $\lceil np_{m}\rceil $ samples with the lowest output scores from the linear model (plus noise) are missing the ``Age'' feature. We generate 200 datasets with MCAR values and 200 datasets with MAR values, and apply both our MCAR test and Little's test to these datasets, and check if these test will reject the hypothesis that the missing is MCAR ($p$ value$<$0.05). Table \ref{tab:mcar} shows the ratio of rejecting null hypothesis. Our method detects MCAR values more reliably than Little's method in high (20\%-30\%) missingness cases.

\textbf{Summary:} We propose an application of EBM to test if the missing value is MCAR. See Appendix~\ref{sec:mcar_case_study} for a case study on infant mortality risk showing that such method may be useful in determining applicability of a model for future data.

% \begin{figure}
%     \centering
%     \includegraphics[width=0.95\linewidth]{Figure/InfantMortalitySmoking.png}  
%     \caption{Impact of smoking before and during pregnancy on infant mortality risk, 2013.}
%     \label{fig:missing_value_not_random1}
% \end{figure}

\subsection{Missing Values Assumed Normal}
\label{sec:missing:normal}
In healthcare domain, it is common for feature values such as lab tests to be missing in the dataset because clinicians believed the patient was likely to be ``normal'' for this measurement, and thus the lab test was not performed \citep{li2021imputation}. In other cases, the measurement may have been made, but the value was not recorded since it was within normal range --- clinicians tend to focus on abnormal findings.

\begin{figure}[h]
    \centering
    % \begin{subfigure}{0.9\linewidth}
    %     \centering
    %     % include first image
    %     \includegraphics[width=1\linewidth]{Figure/missing_assumed_normal_orig.png}  
    %     \caption{Original shape function}
    % \end{subfigure}
    % \begin{subfigure}{0.9\linewidth}
    %     \centering
    %     % include first image
        
        \includegraphics[width=1\linewidth]{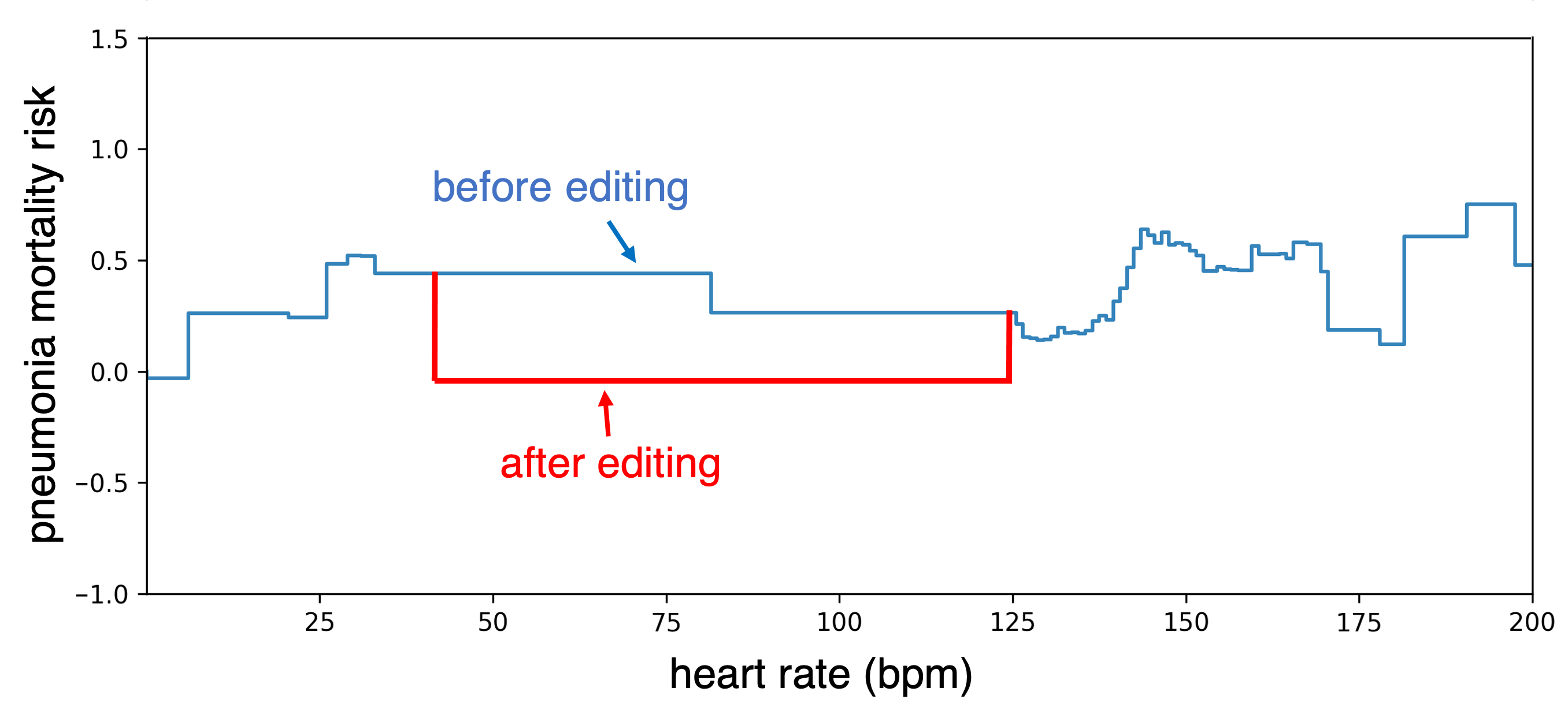}  
    %     \caption{Edited shape function}
    % \end{subfigure}
    \caption{EBM shape function of ``heart rate'' for predicting pneumonia mortality risk. Blue curve is the original shape function; red curve is the edited shape function. }
    \label{fig:missing_value_1}
\end{figure}

For example, this happens to a pneumonia mortality risk dataset \citep{cooper2005predicting}. The blue curve in Figure \ref{fig:missing_value_1} shows what an EBM model has learned for predicting pneumonia mortality as a function of heart rate.
% \note{if this is the first time talking about Pneumonia we may want to briefly summarize it in 2 lines and cite Cooper}
As expected, risk is elevated for patients with abnormally low (10-30) or high heart rate (125-200). The graph, however, shows a surprising region of flat risk between heart rate 38 and 125, which is a normal heart rate for patients in a doctor's office. Moreover, the model surprisingly predicts patients who have \textit{normal} heart rate are at \textit{elevated} risk: it adds 0.22 to the risk for patients in this region.

On further inspection, it turns out that there are no patients in the data set with heart rates between 38 and 125, and 91\% of patients are missing their heart rate which has then been coded as zero. In other words, there are no data to support the model in the normal range of heart rate 38-125, and instead the patients who would be in this range are all coded as zero in the data and on the graph. This explains why the model predicts the lowest risk = -0.04 for patients with heart rate = 0, because these are the patients who actually have normal heart rates.

Any model trained on this data (e.g., boosted trees, random forest, neural networks) is likely to learn to make similar predictions as EBMs in the normal heart rate region because there is no data to support learning the correct risk in this range, and because most models will then interpolate between the extreme regions where they do have data. One exception might be Bayesian models with strong priors, where the prior might dominate in regions of little or no data and cause predictions in this region to be closer to a baseline lower-risk value. The key advantage of using interpretable models such as EBMs is that we can easily see these problems in the model, that ultimately were caused by problems in the data.

If patients with normal heart rates (38-125) will always be coded as zero in the future, then a model trained on this data might always make accurate predictions and the elevated risk predicted by the model in the range 38-125 will not be a problem because no patient will ever fall in that range. However, if the model might be used to make predictions for patients whose true heart rate would be coded between 38 and 125, the model will then make incorrect -- possibly dangerous -- predictions. Thus, it is important to correct this kind of problem. One might hope that a data scientist would detect this kind of problem in the data prior to training a model, however in practice, these kinds of problems can be difficult to detect in the raw data, particularly if there are many different types of problems in the data, and might be easier to detect once an interpretable model is trained. 
(Previous users of the data had not noticed this problem.)

% \note{I wonder if the stuff starting here should be a subsection on model editing}
\subsubsection{Correction via Model Editing}
There are several ways to correct this kind of problem. Of course, the best approach would be to collect and record the true heart rates for all patients. Unfortunately, it is often not possible to go back and correct data in this way. 
% An alternate approach would be to edit the data so that patients coded as zero are randomly assigned heart rates in the interval 38-125, i.e., impute the missing heart rates with a random value selected uniformly from the region where we believe most of the missing values arise from. 
% This, however, does make the assumption that all missing values arise from this one region, and that no patients with low or high heart rate had a missing heart rate. 
% An alternate approach is to use a more sophisticated method of imputing missing values such as random forest imputation \cite{stekhoven2015missforest}.
As we will see in Section \ref{sec:missing:mean}, imputing with the mean or median missing value is probably not ideal. We will also show in Section \ref{sec:missing:advanced} that more advanced methods of imputing missing values such as random forest imputation \citep{stekhoven2015missforest} and KNN imputation \citep{troyanskaya2001missing} might also cause problems.

An alternate approach when EBMs are used is to directly edit the model so that the region 38-125 predicts risk similar to the learned risk prediction for patients with heart rate = 0. Since we do not have any information about true heart rate distribution within the region 38-125, we assume they follow a uniform distribution and edit the graph in this region to be a flat curve. The resulting graph is shown as the red curve in Figure \ref{fig:missing_value_1}. (Note that the result would be similar to uniformly imputing the heart rates in the interval 38-125 and retraining the model.) This approach has the following advantages:
\begin{enumerate}
    \item Editing shape functions provides an opportunity for experts to use their professional training to correct and improve models in ways that may not be adequately represented in the training data.
    \item Editing the model may not only improve the accuracy of the model in the real world where it will be used (instead of just on held-aside test data from the train set), but also make the shape plots more ``reasonable'' and trusted by experts.
    \item Editing an EBM shape function can be done without retraining the model and potentially introducing new problems.
    % \item Correcting the model by editing the data is often much more difficult.
\end{enumerate}

\textbf{Summary:} We show that EBM shape function can help identify the case when feature values are missing because they are assumed to be normal. We also show how editing the EBM graphs can help address issues resulting from missing assumed normal.

\begin{figure}[h]
    \centering
    \begin{subfigure}{1\linewidth}
        \centering
        % include first image
        \includegraphics[width=1\linewidth]{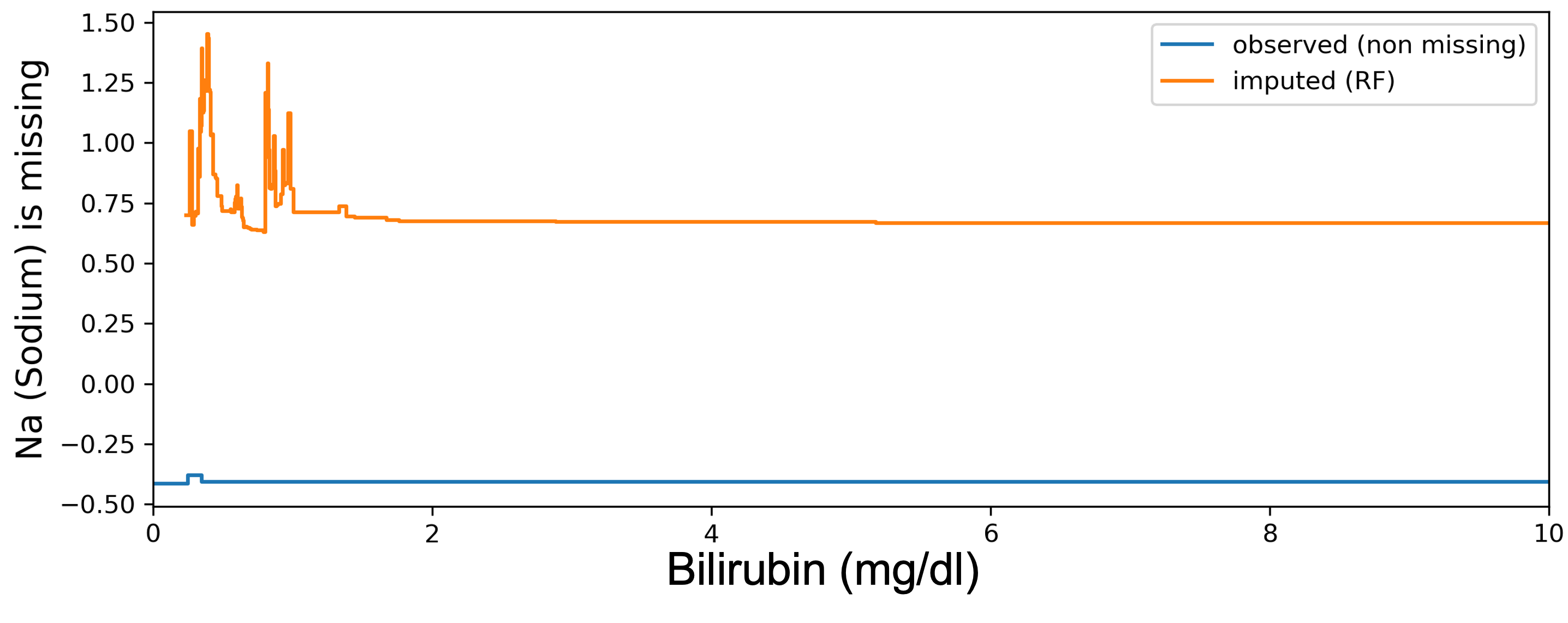}
        \caption{``Bilirubin'' shape function when predicting missingness of ``Na''}
    \end{subfigure}
    \begin{subfigure}{1\linewidth}
        \centering
        % include first image
        \includegraphics[width=1\linewidth]{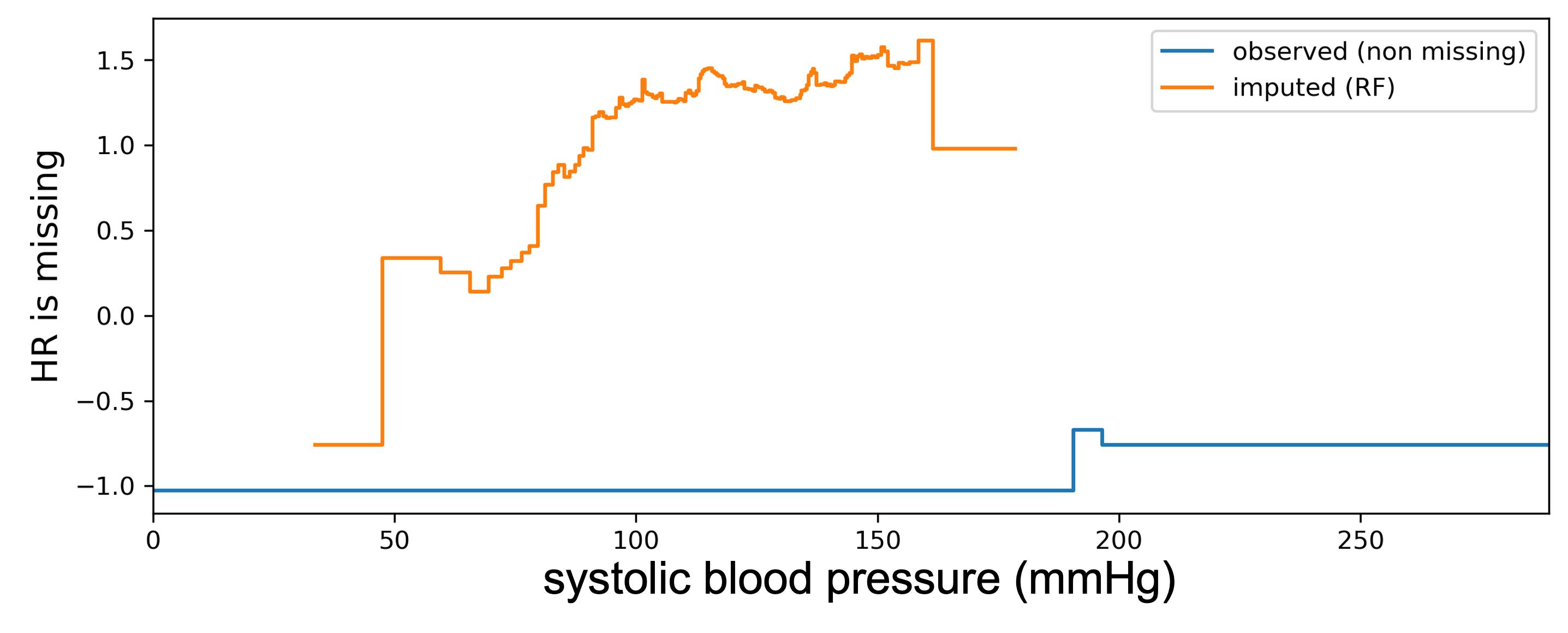}
        \subcaption{``Systolic blood pressure'' shape function when predicting missingness of ``heart rate (HR)''}
    \end{subfigure}
    \begin{subfigure}{1\linewidth}
        \centering
        % include first image
        \includegraphics[width=1\linewidth]{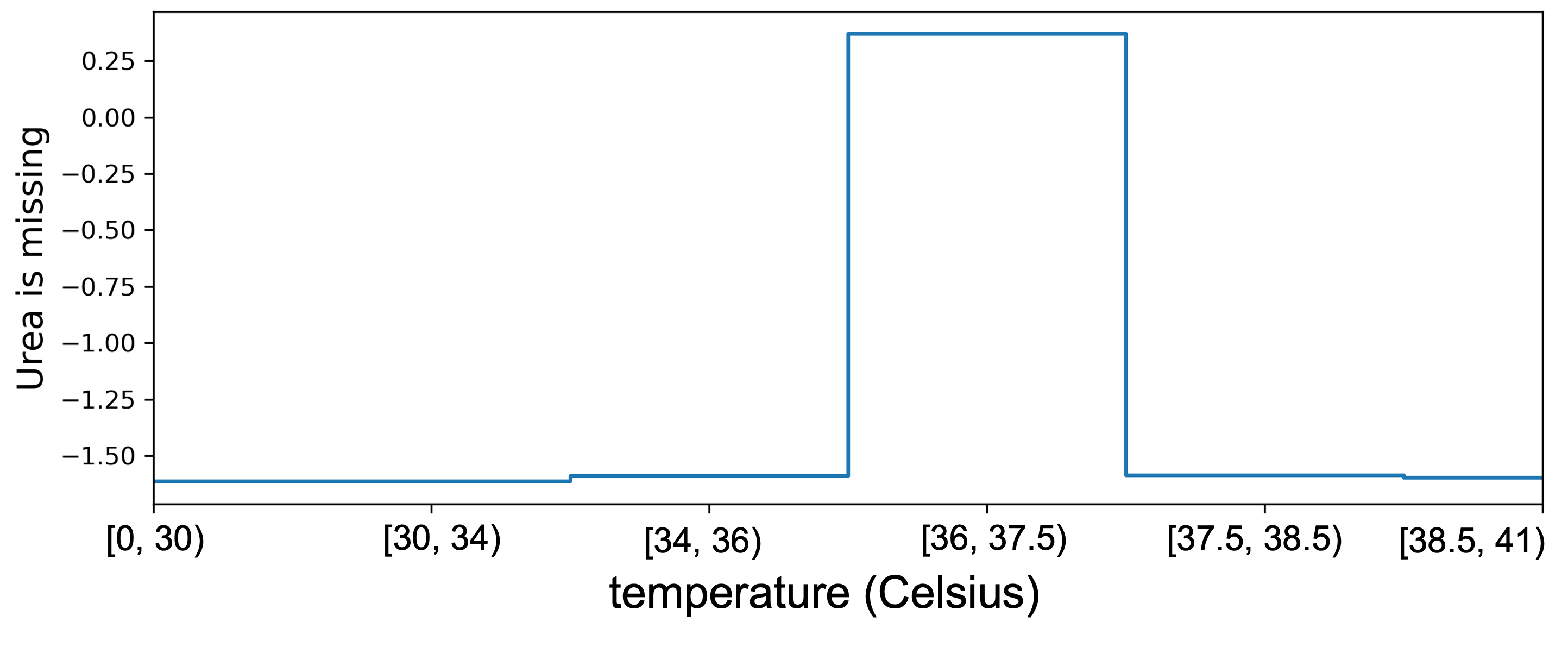}
        \subcaption{``Temperature'' shape function when predicting missingness of  ``Urea''}
    \end{subfigure}
    \caption{EBM shape functions for predicting the missingness of one feature using the others (x-axis: feature value, y-axis: contribution to missingness). The effects of the imputed group (orange) and the non-missing group (blue) are separated.}
    \label{fig:predicting_missingness}
\end{figure}

\begin{table*}[h]
\begin{subtable}[h]{0.47\textwidth}
\scriptsize  %\tiny
\begin{tabular}{|l|l|l|l|l|}
\hline
model & $p_{m}$ & linear & curvilinear & quadratic \\ \hline
LR & \multicolumn{1}{l|}{\multirow{4}{*}{0.1}} & 0.954$\pm$0.014 & 0.902$\pm$0.016 & \textbf{0.883$\pm$0.02}\\ \cline{1-1} \cline{3-5}
RF & & 0.943$\pm$0.014 & 0.946$\pm$0.013 & \textbf{0.883$\pm$0.02} \\ \cline{1-1} \cline{3-5}
KNN & & 0.895$\pm$0.013 & 0.894$\pm$0.009 & 0.881$\pm$0.021\\ \cline{1-1} \cline{3-5}
EBM & & \textbf{0.956$\pm$0.015} & \textbf{0.959$\pm$0.013} & 0.881$\pm$0.02 \\ \hline\hline
 
LR & \multicolumn{1}{l|}{\multirow{4}{*}{0.2}} & 0.928$\pm$0.019 & 0.839$\pm$0.034 & 0.815$\pm$0.013\\ \cline{1-1} \cline{3-5}
RF & & 0.911$\pm$0.019 & 0.928$\pm$0.019 & \textbf{0.831$\pm$0.017} \\ \cline{1-1} \cline{3-5}
KNN & & 0.813$\pm$0.024 & 0.81$\pm$0.022 & 0.812$\pm$0.008\\ \cline{1-1} \cline{3-5}
EBM & & \textbf{0.930$\pm$0.019} & \textbf{0.946$\pm$0.02} & 0.822$\pm$0.016\\ \hline\hline
 
LR & \multicolumn{1}{l|}{\multirow{4}{*}{0.3}} & 0.906$\pm$0.022 & 0.809$\pm$0.054 & 0.710$\pm$0.025\\ \cline{1-1} \cline{3-5}
RF & & 0.887$\pm$0.021 & 0.926$\pm$0.019 & \textbf{0.812$\pm$0.03}\\ \cline{1-1} \cline{3-5}
KNN & & 0.744$\pm$0.032 & 0.752$\pm$0.042 & 0.711$\pm$0.016\\ \cline{1-1} \cline{3-5}
EBM & & \textbf{0.908$\pm$0.022} & \textbf{0.946$\pm$0.02} & 0.795$\pm$0.03\\ \hline
\end{tabular}
\caption{datasets generated by MAR}
\end{subtable}
\vspace{-1mm}
\hfill
\begin{subtable}[h]{0.47\textwidth}
\scriptsize %\tiny
\begin{tabular}{|l|l|l|l|l|}
\hline
model & $p_{m}$ & linear & curvilinear & quadratic \\ \hline
LR & \multicolumn{1}{l|}{\multirow{4}{*}{0.1}} & 0.957$\pm$0.013 & 0.901$\pm$0.013 & \textbf{0.886$\pm$0.017}\\ \cline{1-1} \cline{3-5}
RF & & 0.944$\pm$0.013 & 0.948$\pm$0.011 & \textbf{0.886$\pm$0.017} \\ \cline{1-1} \cline{3-5}
KNN & & 0.899$\pm$0.012 & 0.898$\pm$0.01 & 0.885$\pm$0.018\\ \cline{1-1} \cline{3-5}
EBM & & \textbf{0.959$\pm$0.012} & \textbf{0.963$\pm$0.011} & 0.885$\pm$0.017 \\ \hline\hline
 
LR & \multicolumn{1}{l|}{\multirow{4}{*}{0.2}} & 0.928$\pm$0.018 & 0.847$\pm$0.035 & 0.817$\pm$0.010\\ \cline{1-1} \cline{3-5}
RF & & 0.910$\pm$0.016 & 0.933$\pm$0.016 & \textbf{0.828$\pm$0.012} \\ \cline{1-1} \cline{3-5}
KNN & & 0.816$\pm$0.024 & 0.82$\pm$0.025 & 0.813$\pm$0.008\\ \cline{1-1} \cline{3-5}
EBM & & \textbf{0.931$\pm$0.017} & \textbf{0.953$\pm$0.016} & 0.819$\pm$0.012\\ \hline\hline
 
LR & \multicolumn{1}{l|}{\multirow{4}{*}{0.3}} & 0.914$\pm$0.016 & 0.805$\pm$0.048 & 0.706$\pm$0.024\\ \cline{1-1} \cline{3-5}
RF & & 0.891$\pm$0.015 & 0.925$\pm$0.015 & \textbf{0.811$\pm$0.028}\\ \cline{1-1} \cline{3-5}
KNN & & 0.760$\pm$0.035 & 0.764$\pm$0.039 & 0.711$\pm$0.017\\ \cline{1-1} \cline{3-5}
EBM & & \textbf{0.916$\pm$0.016} & \textbf{0.949$\pm$0.015} & 0.789$\pm$0.03\\ \hline
\end{tabular}
\caption{datasets generated by MNAR}
\end{subtable}
\vspace{-1mm}
\caption{Test accuracy of predicting the missingness. EBM is compared to Logistic Regression (LR), Random Forest (RF), and K Nearest Neighbor(KNN). The accuracies are compared on datasets generated by different missing mechanism (MAR and MNAR generated from linear model, curvilinear model, and quadratic model) with different missing ratio $p_m$ (0.1, 0.2, and 0.3). \label{tab:predict_missing}}
\end{table*}

\subsection{Predicting the Missingness}
Most missing values are not MCAR, but as mentioned in Section \ref{sec:relatedwork}, MNAR and MAR can be difficult to distinguish \citep{van2018flexible}. For both cases, interpretable models like EBM can still be useful in providing insights on possible missingness mechanisms. One way to analyze the missingness mechanism is to predict the missingness of one variable using the other variables (including the target/label). Specifically, the 0-1 missingness indicator is considered as label, and the other features and the label of the original prediction task are considered as input feature to train the machine learning model. The prediction accuracy for missingness tells us roughly how much the missingness is related to the values of other variables. More importantly, with the interpretability of EBMs, we can visualize how the values of these variables contribute to the missingness.

We train EBMs to predict missingness on the MIMIC-II dataset \citep{saeed2002mimic} for every feature that contains missing values. The test AUCs for missingness prediction ranges from 69.10\% to 99.87\% depending on the missing feature: the test AUC is above 84\% for 7 of the 9 missing features. Surprisingly, the test AUC for predicting missingness of ``Na (Sodium)'' and ``Urea'' are 98\% and 99\%, which suggests their missingness can be almost fully explained by other observed variables. Figure \ref{fig:predicting_missingness} shows the shape functions on MIMIC-II, which result from training an EBM on all other variables to predict the missingness of one variable. The features shown in Figure \ref{fig:predicting_missingness} are the features with the largest variable importance for each prediction task. Each shape function shows the contribution of the feature (on the x axis) to the predicted missingness (on the y-axis). Interesting patterns exist in all three graphs and provide insight about why each variables is missing. 

Figure \ref{fig:predicting_missingness}(a) shows how bilirubin contributes to predict the missingness of Na (Sodium). Though bilirubin is a continuous variable and we might expect the shape function to be a continuous curve, the shape function of the observed (non missing) bilirubin samples (in blue) is a constant function with contribution -0.4. This suggests that when bilirubin is measured, Na is less likely to be missing. Moreover, when billirubin is missing and imputed (in orange), there is a large positive contribution (average contribution = +0.93) to the likelihood of Na missingness, which suggests that missing bilirubin strongly predicts that Na will be missing, too.  Interestingly, the causal arrow does not flow the other way: Na is not a strong predictor of bilirubin missingness. The AUC when predicting Na missingness is 0.99, but only 0.73 for predicting bilirubin missingness, and the most important feature for predicting bilirubin missingness is Urea, not Na.  All of this makes clinical sense because bilirubin is included in comprehensive metabolic panels that also always include Na, whereas basic metabolic panels include Na but not bilirubin, which is a more specialized lab test. This also explains why the non-missing group shape function (blue curve) is constant: patients whose Bilirubin are not missing took the comprehensive panels and thus their Na is always measured regardless of the patients' bilirubin value. Remarkably, we are able to detect and understand these effects merely by looking at interpretable EBM models trained to predict missingness. 

We see a similar relationship between heart rate (HR) and blood pressure: when blood pressure is measured, heart rate is almost always measured as well, but it is common to measure heart rate using a finger sensor that does not allow blood pressure to be measured, and this asymmetric relationship between missingness is easily visible by examining EBM plots trained to predict HR missingness. Figure \ref{fig:predicting_missingness}(b) shows the shape functions for observed systolic blood pressure (in blue) and imputed systolic blood pressure (in orange) when predicting whether HR is missing. In the plot, the curve of the imputed group is significantly higher than that of the observed group, again suggesting that when the blood pressure of the patients is missing, their heart rate is also more likely to be missing. This effect is strong, as the maximum gap between the two curves is approximately 2.5 (1.5 in orange curve and -1.0 in blue curve) of predicted log odds. Again the blue curve is constant.

Figure \ref{fig:predicting_missingness}(c) shows the shape function for temperature when predicting if urea is missing or not. There is no missing value for temperature, so there is no orange curve. The bump at temperature $\in$ $[36, 37.5)$ indicates that urea is more likely to be missing, which suggests when a patient has normal body temperature, doctors may be less likely to order a blood test to measure urea.

To test how well can EBM predict the missingness, we generate some semi-synthetic datasets with ground-truth missing mechanism. Again, these semi-synthetic datasets start from MIMIC-II imputed by MissForest, and then apply fixed models (linear, curvilinear and quadratic models) plus an Gaussian noise to decide which entry in the ``Age'' feature is missing. The feature value is missing when the output score is higher than the threshold. The difference between MAR and MNAR is whether the target feature value is considered as an input of the missing models. Table \ref{tab:predict_missing} compares EBM's the test accuracy of predicting missingness with machine learning models commonly used for missing value imputation. EBM predicts missingness better than other methods in cases of MAR and MNAR values generated from linear and curvilinear models and is not far behind Random Forest in case of quadratic model. 

\textbf{Summary:} We use EBMs to  predict the missingness of features from other input features. EBM predicts the missingness accurately. The interpretability of EBMs can help users understand the relationship between the features and missingness and thus bring more insight for the cause(s) of missingness.

\section{Detecting and avoiding potential risks of missing value imputations}
\label{sec:imputation}
\subsection{Imputation With the Mean}
\label{sec:missing:mean}
Because many machine learning methods cannot natively handle missing values, it is common for data scientists to impute missing values before training models. There are many different ways to %impute missing values 
do this
\citep{lin2020missing}: with the mean, the median, with a unique value such as 0 or -99 or +99, or by using a machine learning method such as MissForest.
% See \cite{lin2020missing} for an overview of imputation methods.  

%\begin{figure}
%    \centering
%    \includegraphics[width=.95\linewidth]{Figure/mean_imputation_pfratio.png}
%    \caption{EBM shape function and density plot for P/F ratio when predicting ICU mortality.}
%    \label{fig:missing_value_2}
% \end{figure}

Perhaps the most common form of missing value imputation is to use the mean, but this can sometimes be problematic. Figure \ref{fig:missing_value_2} shows an EBM plot of the mortality risk of ICU patients as a function of their P/F ratio.  P/F ratio is a measure of how well a patient converts oxygen in the air they breathe into oxygen in their blood: low P/F ratio indicates patients with low blood-oxygen whose lung function is impaired, while P/F ratio around 1000 and higher indicates good lung function.  As expected, the learned shape function captures this.  What is surprising, however, is the large drop in risk at about P/F ratio=323.  What could cause that?

A simple test for blood-oxygen levels is to pinch a fingertip and see how quickly color returns to the skin.  If color returns quickly, clinicians know the blood-oxygen level is good and do not bother to measure P/F ratio --- the P/F ratio is assumed normal.  In this dataset, however, the missing P/F ratio values were imputed with the mean instead of being coded as 0 as they were in Figure \ref{fig:missing_value_1}. 60\% of patients are missing P/F ratio. The mean P/F ratio when not missing (40\% of the data) is 323.6, so 60\% of patients have had their P/F ratio imputed with this value. Because this is a large sample of healthy patients with strong respiration, the model learns that their risk is comparable to the risk of other healthy patients with P/F ratio above 1000. This explains why the graph dips at 323, yet predicts higher risk just before and after this value. Although this anomaly does not significantly hurt the accuracy of the model because it has learned to make appropriate low-risk predictions for the 60\% of patients at this value, it is risky to leave this anomaly in the model because there are real patients with P/F ratio$\approx$323 who will be predicted to have low risk but who are genuinely at elevated risk. For this reason, it would be better to 
% either use a more sophisticated method of imputing missing values, or to 
encode the missing value with unique value (e.g., -1). Model editing is not a good solution for this problem because imputation with the mean has caused patients who are low risk (missing values) and elevated risk (P/F ratio near 323) to fall at the same place on the shape function, thus there is no reasonable edit to the graph that can predict the correct risk for both groups.

\subsubsection{Automatic Detection of Bad Imputations}
\begin{figure}[h]
    \centering
    \begin{subfigure}{1\linewidth}
        \centering
        % include first image
        \includegraphics[width=1\linewidth]{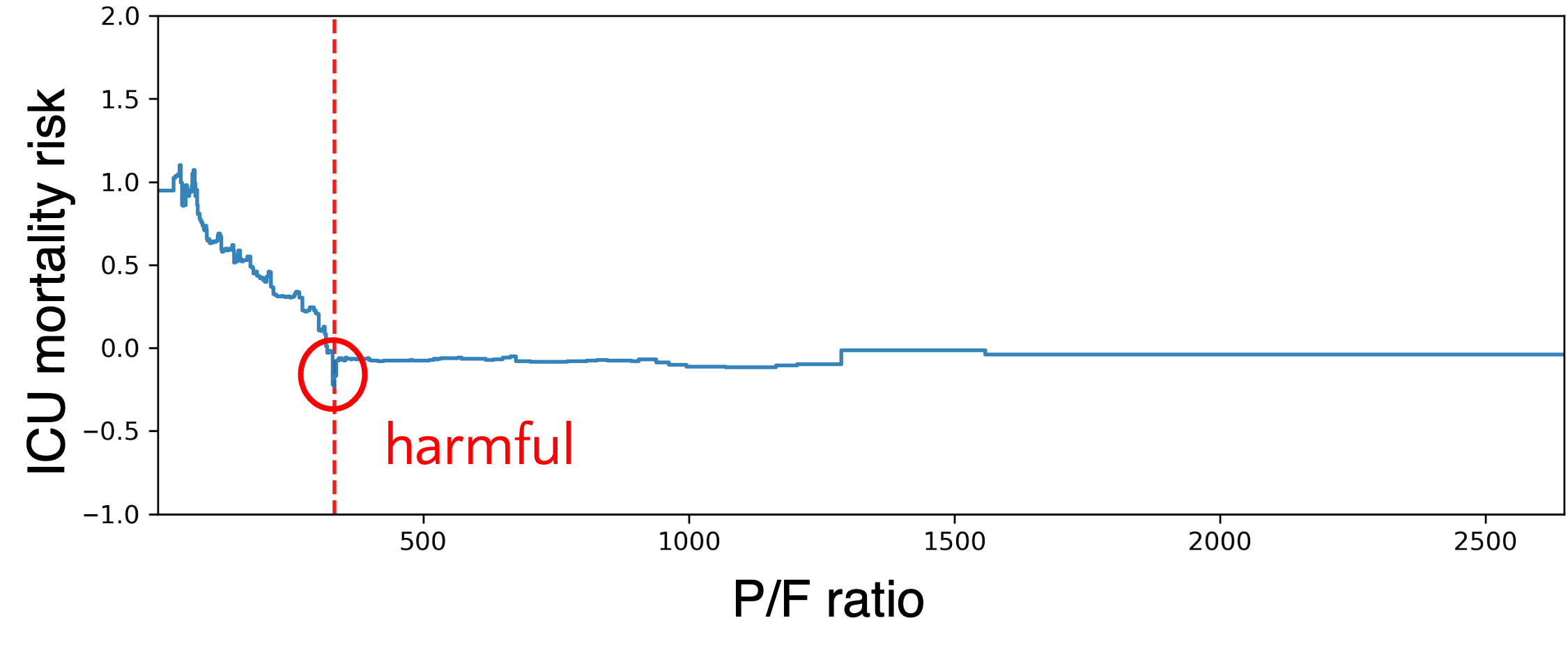}  
        \caption{Shape function for P/F ratio.}
    \end{subfigure}
    \begin{subfigure}{1\linewidth}
        \centering
        % include first image
        \includegraphics[width=1\linewidth]{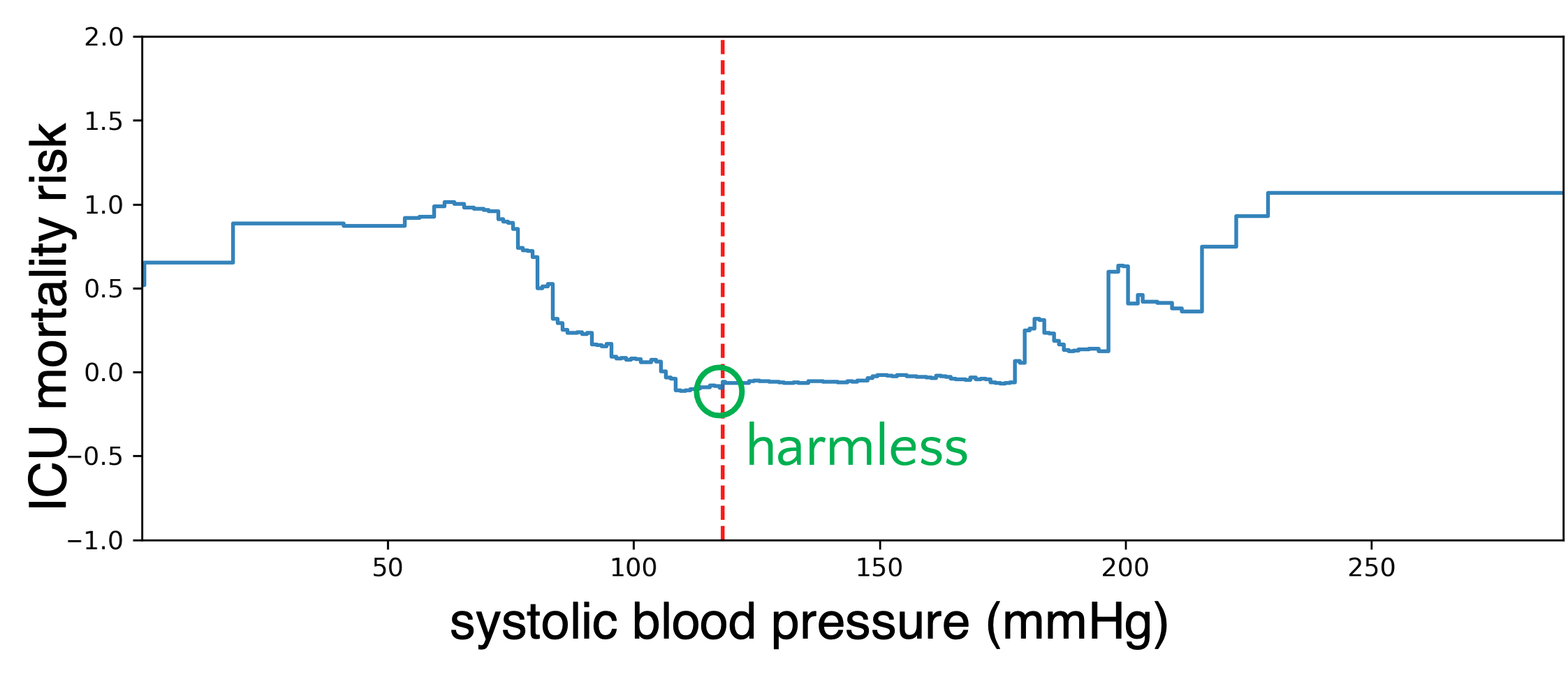}
        \caption{Shape function for systolic blood pressure.}
    \end{subfigure}
    \begin{subfigure}{1\linewidth}
        \centering
        % include first image
        \includegraphics[width=1\linewidth]{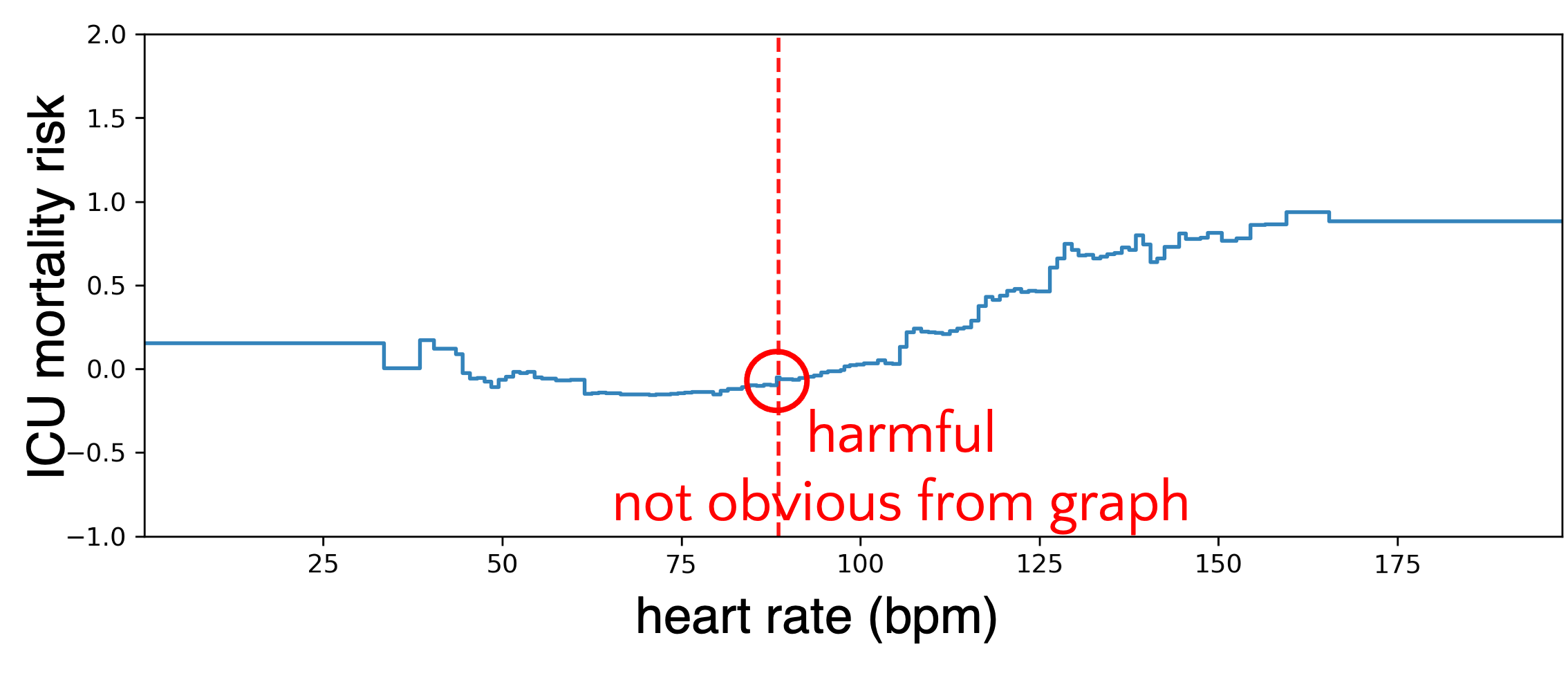}
        \caption{Shape function for heart rate.}
    \end{subfigure}
    \caption{Examples of potentially harmful (a) \& (c) and harmless mean imputations (b) found by our automatic detection algorithm. Red vertical lines indicate average feature values. A large spike at the mean is potentially harmful while a small spike or no spike is harmless.}
    \label{fig:missing_value_detection}
    \vspace{-2mm}
\end{figure}

As discussed in the P/F ratio example above, mean imputation could be dangerous especially when the missing group is significantly different from the samples with feature values near the mean. As shown in Figure \ref{fig:missing_value_2} and Figure \ref{fig:missing_value_detection}(a), such distribution differences can be reflected as spikes on the EBM shape functions. However, if the spike is small or there is no spike near the mean value, e.g., Figure \ref{fig:missing_value_detection} (b), the difference between groups might be insignificant and mean imputation can be harmless. Since bad mean imputation is associated with the spikes at the mean, can we automatically detect bad mean imputations through a spike detector? The answer is yes, but we need to address two problems, (1) how to know the spike is at the mean (2) how to detect spikes, given that the shape function itself can fluctuate. 

The first problem is easy to solve. Observing that the mean value of the feature is the same before and after mean imputation, we can directly find the bin (of EBM) covering the mean value, and detect if the bin is a spike or not. This also works for median imputation --- the median of a feature does not change by imputing the missing values with the median.

To address the second problem, we need an algorithm to distinguish spikes resulting from mean imputation and fluctuations that naturally occur in the EBM shape functions. We formulate this as an outlier detection problem. First, we calculate the second order differences for all bins in all shape functions (excluding first and last bins), since spikes usually have extreme second order differences. We denote the function values of the $k^{th}$ bin and its neighbouring bins as $f_k$, $f_{k-1}$, and $f_{k+1}$. The corresponding bin sizes are denoted as $h_k$, $h_{k-1}$, and $h_{k+1}$. The second order difference is
\begin{equation}
    f''_k(x) \approx \frac{\frac{f_{k+1}-f_{k}}{(h_{k+1}+h_{k})/2} - \frac{f_{k}-f_{k-1}}{(h_{k}+h_{k-1})/2}}{h_k+h_{k+1}/2+h_{k-1}/2}.
\end{equation}
We then run an outlier detection algorithm (Isolation Forest \citep{liu2008isolation}) on these second order differences. The algorithm predicts an anomaly score for each bin, and we choose a threshold so that around 5\% of bins are detected as outliers. The potentially harmful mean imputations are predicted if bins covering the mean values are also predicted as outliers. The same procedure is also applied to detect potentially harmful median imputations.

We test the bad mean imputation detection algorithm on the MIMIC-II dataset with mean imputation on continuous features. Among the 13 continuous features, in 4 a spike is detected at the mean. Other continuous features do not have a spike at the mean and are predicted to be ``harmless'' in terms of mean imputation. As expected, continuous features with \textit{no} missing values are predicted as negative. Figure \ref{fig:missing_value_detection}(c) shows a potentially harmful mean imputation found by our detection algorithm but not discovered visually as the spike is not obvious. This represents one of the smallest spikes that the anomaly detection algorithm would detect as potentially harmful mean imputation (given this sample size).

\textbf{Summary:} By examining anomalies in the EBM shape functions one can easily identify bad imputations with the mean. Based on this finding, we propose an automatic detection method to detect imputations with the mean that can be potentially risky.

\subsection{Imputation With Advanced Methods}
\label{sec:missing:advanced}

\begin{figure}[h]
    \centering
    \begin{subfigure}{1\linewidth}
        \centering
        % include first image
        \includegraphics[width=1\linewidth]{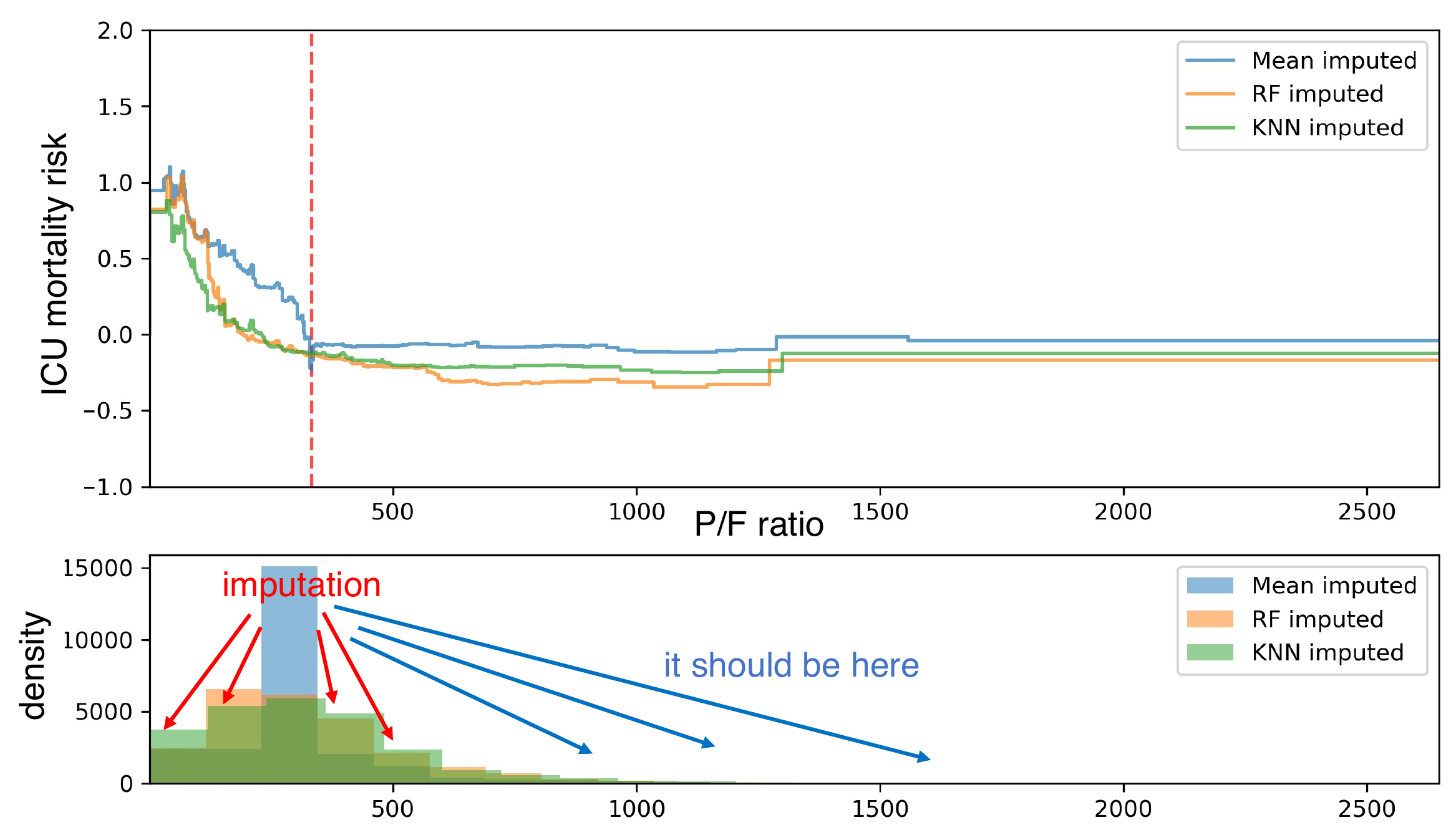}
        \caption{Shape functions and density plots for P/F ratio}
    \end{subfigure}
    \begin{subfigure}{1\linewidth}
        \centering
        % include first image
        \includegraphics[width=1\linewidth]{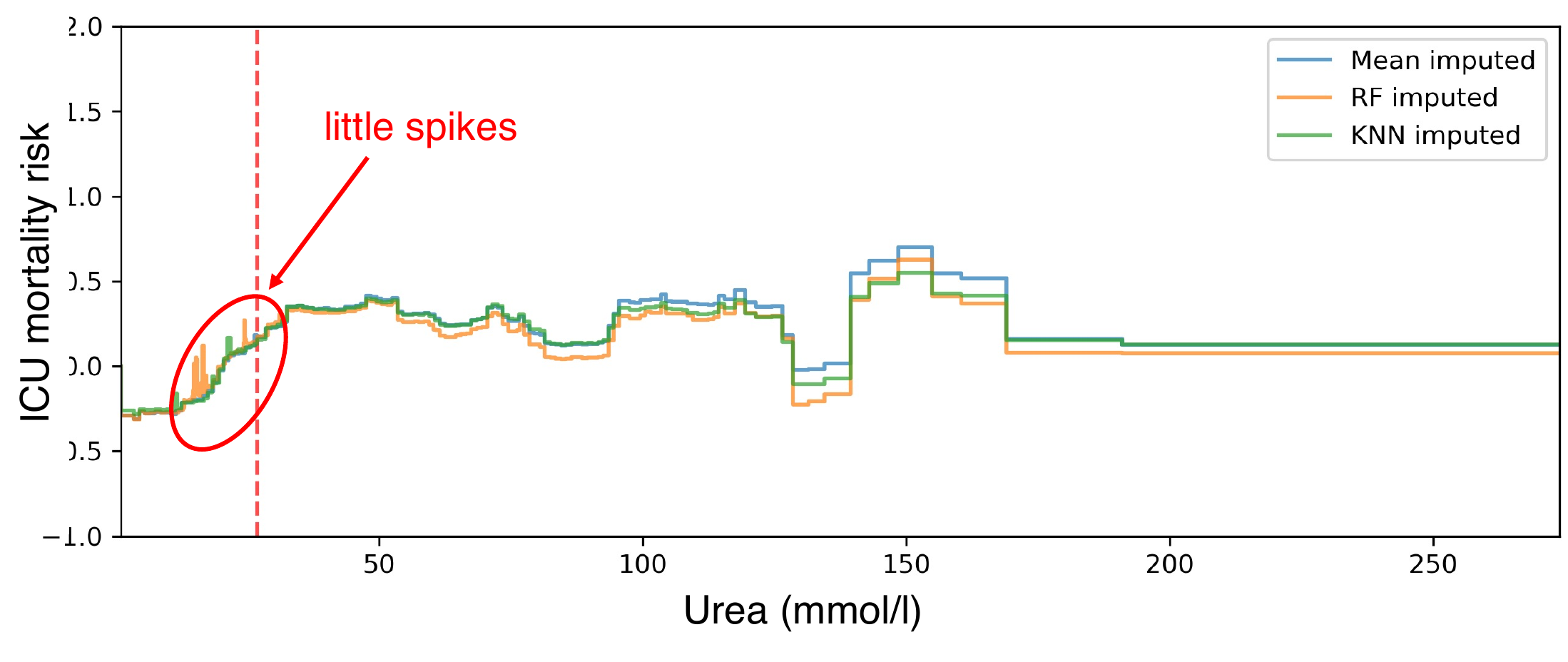}
        \caption{Shape functions for Urea}
    \end{subfigure}
    \vspace{-1mm}
    \caption{EBM shape functions trained on datasets imputed with different missing value imputation methods (mean imputation, MissForest (RF) imputation and KNN imputation).}
    \label{fig:missing_value_3}
\end{figure}

One might assume that imputation with more advanced methods such as MissForest (RF) imputation \citep{stekhoven2015missforest} or k-nearest neighbor (KNN) imputation \citep{troyanskaya2001missing} would not exhibit problems like those discussed in Sections \ref{sec:missing:normal} and \ref{sec:missing:mean}, because they are designed to impute feature values based on the conditional feature distribution in the data. For example, MissForest iteratively trains a random forest regression model, predicting and updating the missing values of each covariate using the other covariates, until these values converge. %However, i
%In this section we will show how 
Interpretable EBM models can help us detect unexpected problems that can be caused by imputation with these methods.

We apply different imputation methods (mean imputation, MissForest (RF) and KNN imputation) to the MIMIC-II dataset. Figure \ref{fig:missing_value_3}(a) shows the P/F ratio shape functions and densities for the different imputation methods. As described in Section \ref{sec:missing:mean}, values near 1000 are healthy, and lower P/F ratio indicates poor lung function. In this dataset, P/F ratios are missing when doctors assume they are normal, i.e., the ground truth of missing values are likely to be near 1000. However, the density plot shows that instead of imputing missing values with P/F ratio values near 1000, RF and KNN actually impute P/F ratio with lower values. Such imputations are problematic because they systematically reduce the predicted risk of the riskier low-P/F-ratio patients and those patients might then not receive adequate care if the resulting model is used clinically. Compared with mean imputation, the advanced imputation methods actually affect a larger range of patients (P/F ratio between 0 and 800) and the advanced methods could be even more harmful than  mean imputation.

Another problem of advanced imputation methods is that they can sometimes introduce fluctuations to models which show up as little spikes on EBM shape functions. Figure \ref{fig:missing_value_3}(b) shows the EBM shape function for the feature ``Urea'' with many little spikes when the missing values are imputed with RF and KNN. Again, such spikes can be potentially harmful for patients with almost the same feature values at these locations. The fluctuation problem can be resolved if the model enforces local smoothness (e.g., linear models or GAMs with smooth splines). However, tree-based models like random forests, gradient boosted trees, and EBMs often are not locally smooth and are likely to learn such spikes.

\textbf{Summary:} We show that advanced imputation methods like MissForest and KNN %imputations 
can create problems for machine learning models that are hard to detect. We propose a way to use EBMs to visualize the potential impact of these imputation methods (Appendix~\ref{sec:visualize_imputation}), and show that it helps detect potential problems that otherwise might have remained invisible and led to suboptimal healthcare decisions.

% However, since we do not know the ground truth of these missing HRs in pneumonia dataset, we cannot evaluate the test performance of the edited model. To test the validity of such editing, we create a semi-simulated dataset with ground truth of the  missing values based on the MIMIC-II dataset. We chose MIMIC-II because it also has HR feature but the HR values are not missing. Specifically, we manually create the missing values by setting all HR values between 40 and 110 to 0. We first train an EBM on the original training set without missing HR, and test it on the original test set. The test accuracy is 83.90\%. We then train another EBM on the training set with missing HR, and test it on both the test set with missing HR and the \textit{original} test set.

\section{Discussion}
\label{sec:discussion}
% \subsection{Small Edits to Models Often Yield Small Changes in Accuracy but Can Still be Important}
We found many potential risks in models that were introduced by missing values or imputation. Because EBMs are interpretable and editable, once the problem is detected, we can often edit the model to fix these issues using existing model editing tools for GAMs \citep{wang2021gam}. Because edits only affect model behavior on small subsets of samples and for a few features (e.g., samples near the mean in the case of mean imputation), the change in accuracy is small. However, these changes can still be critical in high-stakes tasks like medical care, where the potential cost for bad predictions is very high.

The proposed methods are all based on EBM. We chose EBMs because the shape functions are good at capturing subtle anomalies in the data, compared to linear models and decision trees. In the future, it is worth investigating if other interpretability methods can handle the same missing value tasks. For example, a sparse decision tree model \citep{lin2020generalized} might be able to learn complex feature interactions when predicting missingness from other features.

\section{Conclusion}

\label{sec:conclusion}
% \note{there are some very  interesting points here, but a bit speculative. I wonder if it would be better as a discussion section and then a very short conclusion section that repeats our contribution again}
We propose methods based on glass-box EBMs to help understand and address missing value problems. Such problems are common in medical applications.  Experiments on real-world medical datasets show that the proposed methods provide insights on the causes of missingness, and can also help detect and avoid potential risks introduced by different imputation methods. Specifically, in terms of understanding missingness, we propose a novel method using EBMs to test for MCAR. For the non-MCAR case, we show that EBM shape functions can help identify when feature values are missing because they were assumed to be in the normal range for that variable. We also use EBMs to predict the missingness of some features from other input features. Here the interpretability of the model can help users better understand the relationship between features and missingness. For imputation, we show that anomalies in the EBM shape functions can be used to automatically identify potentially harmful imputation with the mean or median. For advanced methods like MissForest and KNN imputation, we propose methods for visualizing the potential impact of imputation on the resulting model. 

\clearpage
\bibliography{mybibliography}

\appendix
\clearpage

\section{Testing for MCAR with EBM: Case Study}
\label{sec:mcar_case_study}

In some cases we have information about the mechanism generating missing values and the likelihood that a similar mechanism will generate data in the future.  

As an example, consider CDC Birth Cohort Linked Birth – Infant Death Data Files \cite{CDC:InfantLinkedDatasets}.  The dataset describes pregnancy and birth variables for all live births in the U.S. together with an indication of an infant's death before the first birthday.  The dataset is collected using two certificates:  1989 Revision of the U.S. Standard Certificate of Live Birth (unrevised) and the 2003 revision of the U.S. Standard Certificate of Live Birth (revised).  As a result of the delayed, phased transition to the 2003 Certificate, the cohorts from 2004 to 2015 include data for reporting areas that use the newer 2003 revision along with data for reporting areas that still use the older 1989 Certificate (unrevised), with later years having a larger fraction of data corresponding to the 2003 revision. Values for variables that are present only in the 2003 certificate will be missing for areas using the earlier, 1989 certificate.  In 2013, 10\% of records come from such areas, the fraction is declining year to year and we can expect it to be even smaller in subsequent years.

\begin{figure}
    \centering
    \includegraphics[width=0.95\linewidth]{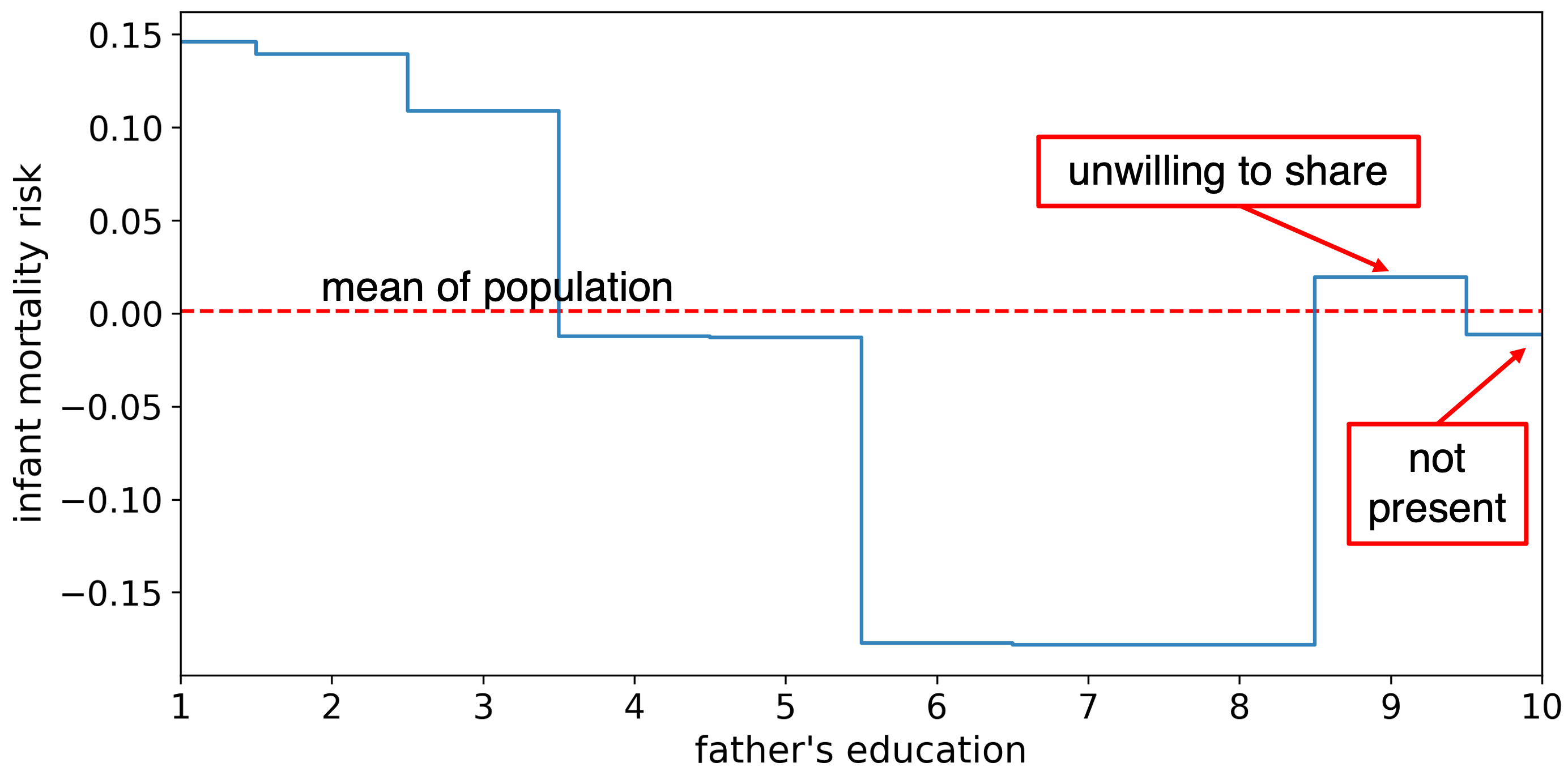}  
    \caption{Impact of father's education on infant mortality risk, 2013.}
    \label{fig:missing_value_not_random}
\end{figure}

\begin{figure}
    \centering
    \includegraphics[width=0.95\linewidth]{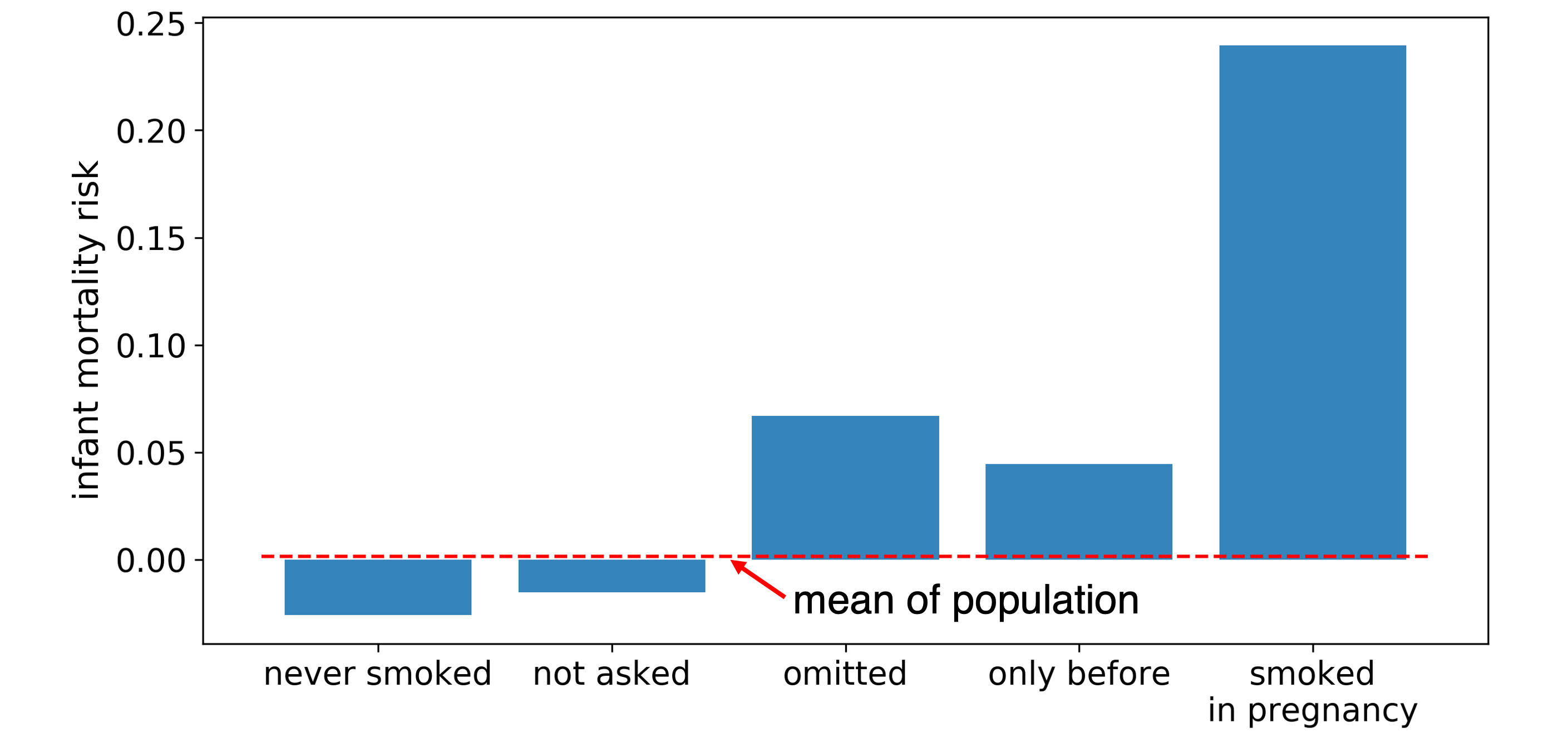}  
    \caption{Impact of smoking before and during pregnancy on infant mortality risk, 2013.}
    \label{fig:missing_value_not_random1}
\end{figure}

Figure \ref{fig:missing_value_not_random} shows the impact of father's education on infant mortality risk according to an EBM model trained on 2013 data.  Values from 1 to 8 correspond to different levels of educational attainment, with 1 indicating 8th grade or less and 8 a doctorate or professional degree. The risk is high for levels 
%1 through 3, 
1-3, 
drops to just below the average risk for levels 
%3 and 4
3-4 
(some college and associate degree) and even further for BA/BS, MA/MS and doctorate (levels 6-8)\footnote{Parents' education is the best proxy we have in the dataset for family's income.}. Level 9 indicates 
%the parents not willing 
unwillingness to share this information and 10 corresponds to 10\% of records where this variable was not present (version 1989).
%of the certificate
Level 9 is associated with slightly elevated risk; we may guess that fathers unwilling to share are more likely to be lower on the education scale.  Level 10 is associated with risk slightly below average, which is surprising at first glance.  Unlike for Level 9, the mechanism according to which the information is withheld is independent of the value of the variable in question (namely, the geographical area using an older version of the certificate).  However, if the populations using the two certificate versions were coming from the same distribution, we would expect average risk (0 on the shape function) for this group.  The MCAR test from Section \ref{sec:mcar} indicates these groups are statistically different from each other, suggesting social, demographic %cultural, 
or other differences between these populations. 

A similar picture emerges when we look at infant mortality as a function of mother smoking before and during pregnancy. The risk is highest for mothers who smoked during pregnancy, slightly elevated for those who smoked before pregnancy and lowest for mothers who never smoked.  Risk for mothers who didn't share this information (`omitted') is clearly elevated. The group for whom the value is missing (older 1989 certificate, denoted `not asked') has risk slightly lower than average (0). Again, risk different from average indicates a distribution shift with respect to the rest of the population, and we see that 'omitted' is different from 'not asked'.

If we were to train an infant mortality risk model on 2013 data and use it for prediction on data from subsequent years, we could run into the problem of values missing for an even lower fraction of all records and possibly coming from a distribution even more shifted with respect to the distribution of the majority of the records.  Our model would likely predict the risk less accurately for this segment of the population.

\section{Visualizing the Effect of Imputation}
\label{sec:visualize_imputation}
As mentioned in Section~\ref{sec:imputation}, the advanced imputation methods can significantly change the learned shape functions and such changes can sometimes be problematic. To help visualize the effect of imputation and identify potential problems in advance, we propose to separate the components of the missing group and the observed group in the EBM shape functions. To separate these two components, instead of directly imputing the missing values with the output of the imputation algorithm, we add a large offset to these imputed values so that the imputed values do not have overlap with the observed values. For example, in our experiments, we add max feature value plus 1 to the imputed values. This can be viewed as a trick to squeeze the feature and its missingness indicator variable into one dimension. Training EBMs on such separated feature values, the shape function will be a concatenation of the two curves corresponding to the observed group and the missing group. Also, because we know the offset we added to the imputed value, we can subtract it during visualization, and show the two curves on the same plot and original x-axis. 

\begin{figure}[h]
    \centering
    \begin{subfigure}{1\linewidth}
        \centering
        % include first image
        \includegraphics[width=1\linewidth]{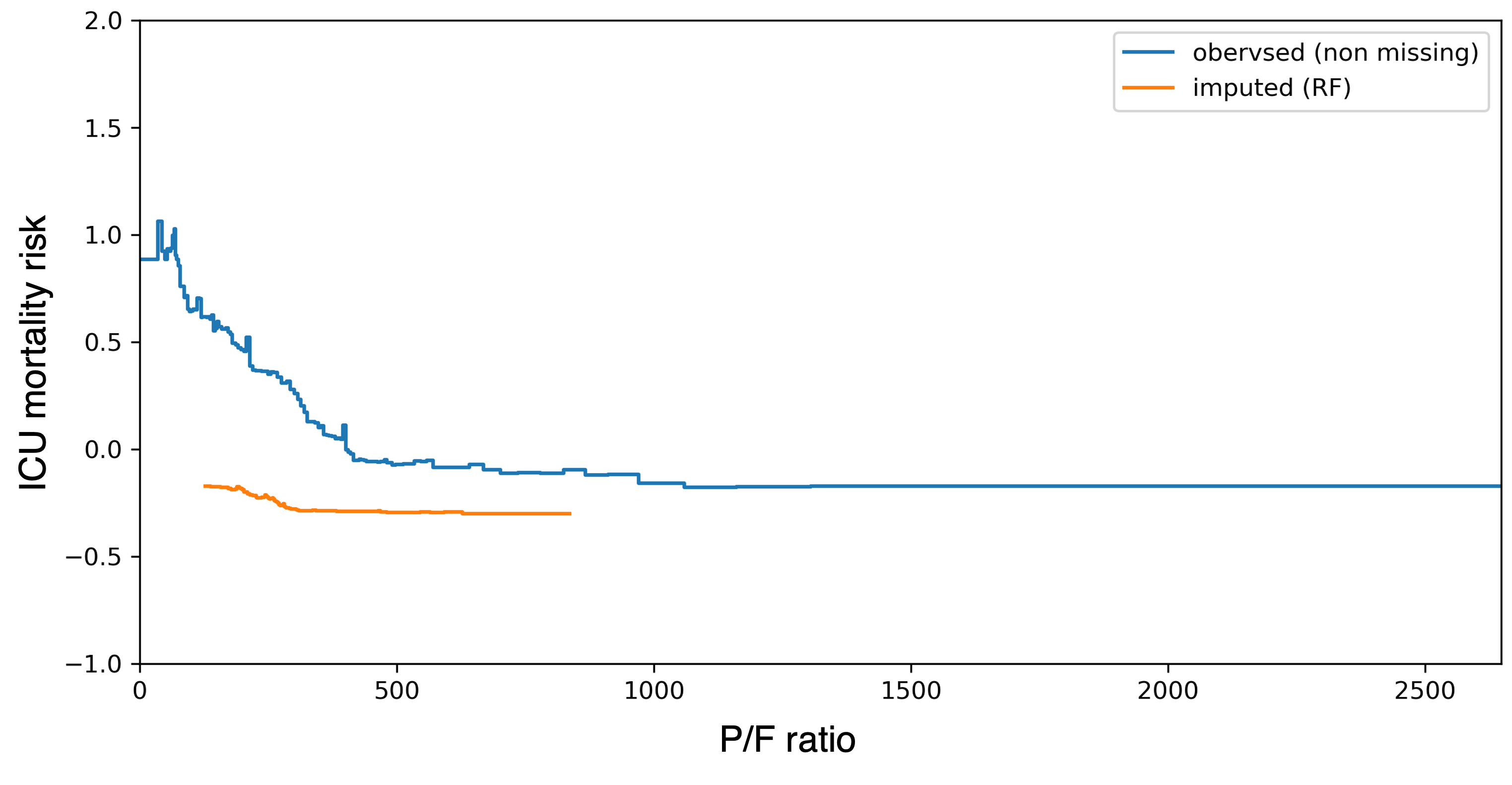}
        \caption{Shape functions for P/F ratio}
    \end{subfigure}
    \begin{subfigure}{1\linewidth}
        \centering
        % include first image
        \includegraphics[width=1\linewidth]{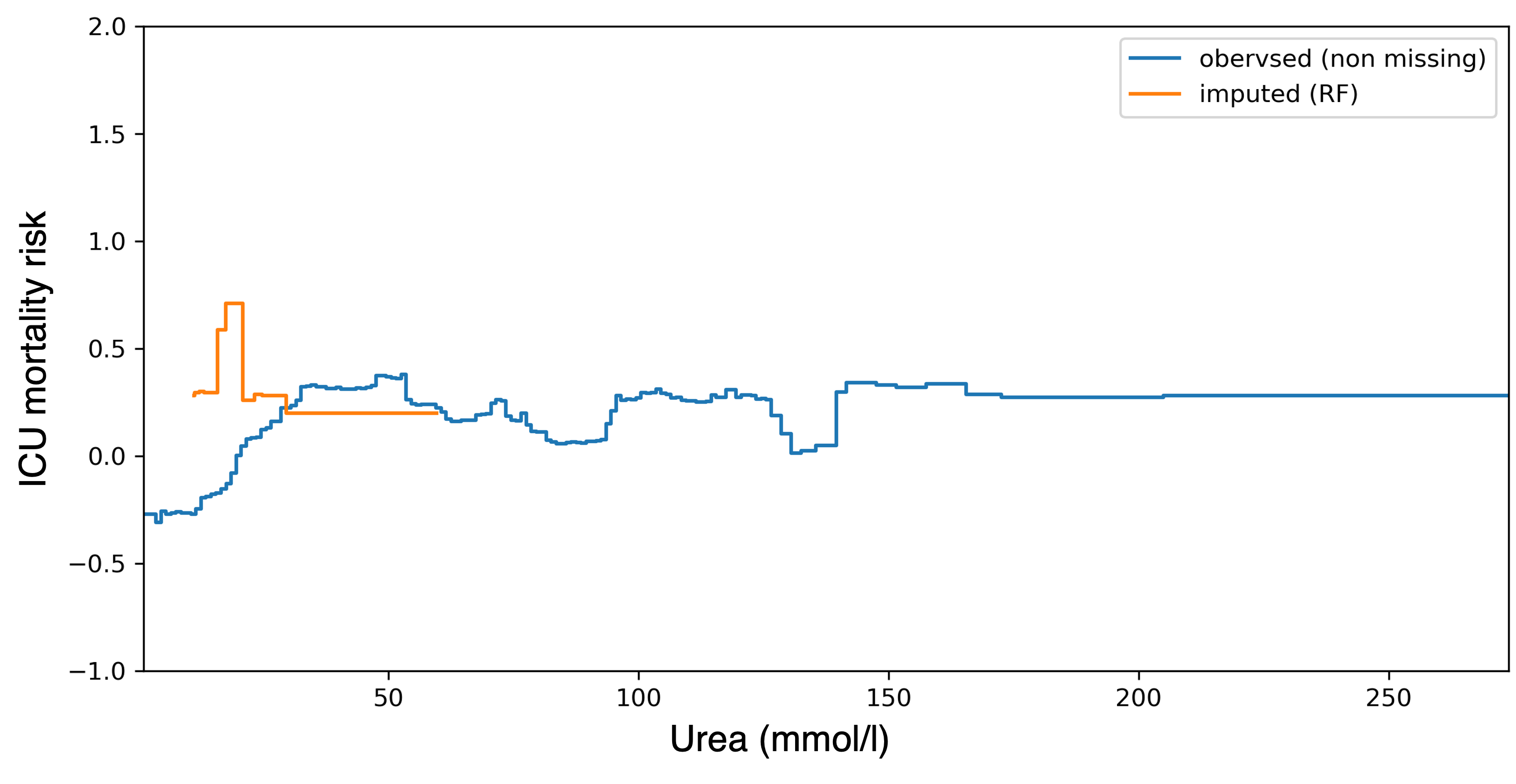}
        \caption{Shape functions for Urea}
    \end{subfigure}
    \caption{EBM shape functions when the effects of imputation group (imputed by MissForest, denoted as RF imputed) and observed (non missing) groups are separated. The plots suggests how the two groups are different in terms of predicting the ICU mortality risk, and suggests how MissForest imputation might result in problematic models.}
    \label{fig:imputation_effect}
\end{figure}

Figure \ref{fig:imputation_effect} shows the EBM shape functions of the imputed group and the observed group separated using the method proposed above. Figure \ref{fig:imputation_effect}(a) shows that the risk of the RF imputed group is much lower than the risk of the observed group which corroborates what we found in Figure \ref{fig:missing_value_3}(a). Similarly, the effects of the imputed group in Figure \ref{fig:imputation_effect}(b) also differ significantly from the observed group, which explains why there exist spikes in the RF imputed EBM shape function in Figure \ref{fig:missing_value_3}(b). Using interpretable methods like EBMs allows one to understand the consequence of different imputation methods that otherwise would be invisible.

\end{document}